\documentclass[runningheads]{llncs}

\usepackage[T1]{fontenc}
\usepackage{graphicx}
\usepackage{amsmath,amssymb}
\usepackage{booktabs}
\usepackage{array}
\usepackage{multicol}
\usepackage[pass]{geometry}
\usepackage[table]{xcolor}
\usepackage{url}
\usepackage[hidelinks]{hyperref}
\newcommand{\verySmallP}{\ensuremath{\ll}}
\newcommand{\method}{SegWithU}
\newcommand{\Rset}{\mathbb{R}}
\newcommand{\indicator}{\mathbf{1}}
\begin{document}

\title{\method: Deterministic Perturbation Probes for Single-Backbone-Pass Risk-Aware Medical Image Segmentation}
\titlerunning{\method{} for Risk-Aware Medical Segmentation}
\author{
\href{https://orcid.org/0009-0007-2342-5350}{\includegraphics[scale=0.06]{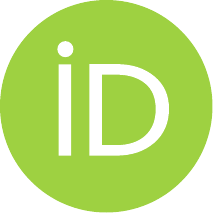}\hspace{1mm}Tianhao Fu} \textsuperscript{1, 4, 5, \thanks{Project lead with correspondence at \href{mailto:tianhao.fu@mail.utoronto.ca}{tianhao.fu@mail.utoronto.ca}.}}\and
\href{https://orcid.org/0009-0008-3817-0589}{\includegraphics[scale=0.06]{images/orcid.pdf}\hspace{1mm}Austin Wang} $^\dagger$\textsuperscript{, 2, 5}\and
\href{https://orcid.org/0009-0005-7789-3876}{\includegraphics[scale=0.06]{images/orcid.pdf}\hspace{1mm}Charles Chen} $^\dagger$\textsuperscript{, 1, 5}\and
\href{https://orcid.org/0009-0008-0149-2250}{\includegraphics[scale=0.06]{images/orcid.pdf}\hspace{1mm}Roby Aldave-Garza} $^\dagger$\textsuperscript{, 3, 5}\and
\href{https://orcid.org/0009-0000-9492-8958}{\includegraphics[scale=0.06]{images/orcid.pdf}\hspace{1mm}Yucheng Chen} \textsuperscript{5}
}
\authorrunning{T. Fu, A. Wang, C. Chen, R. Garza, Y. Chen}
\institute{
\small$^\dagger$ Equal contributions.\\
\small\textsuperscript{1} University of Toronto, Toronto, ON, Canada\\
\small\textsuperscript{2} McGill University, Montreal, QC, Canada\\
\small\textsuperscript{3} University of Waterloo, Waterloo, ON, Canada\\
\small\textsuperscript{4} Vector Institute, Toronto, ON, Canada\\
\small\textsuperscript{5} Project Neura, Toronto, ON, Canada\\
}

\maketitle

\begin{abstract}
Reliable uncertainty maps are needed when medical segmentations drive downstream measurements, triage, or clinical review. Strong uncertainty baselines often require repeated inference, while efficient single-backbone-pass methods can be weak at ranking failures or depend on restrictive feature-space assumptions. We present \method, a supervised frozen-backbone uncertainty module for medical image segmentation. \method{} leaves a pretrained 3D segmentation backbone unchanged, taps decoder features, and learns a compact probe space whose deterministic signed perturbation patterns measure local prediction instability. The module produces two voxel-wise maps with distinct roles: a calibration map for probability tempering and a ranking map for error localization and selective prediction. We explicitly formulate the perturbation patterns, map dependencies, training losses, and voxel-level evaluation protocol. On ACDC, BraTS2024, and LiTS, \method{} achieves AUROC/AURC of $0.9838/2.4885$, $0.9946/0.2660$, and $0.9925/0.8193$, respectively, with AURC reported in $10^{-4}$ units, while preserving the frozen backbone's hard segmentations. Ablations show that the calibration/ranking decomposition, ranking losses, and learned perturbation probes each contribute to performance.
Code will be released after review.
\end{abstract}

\section{Introduction}
\label{sec:introduction}

Medical image segmentation is increasingly used as a quantitative instrument for anatomical measurement, lesion burden estimation, treatment planning, and longitudinal monitoring. Because plausible contours can still contain clinically relevant errors, a deployed segmentor should expose not only what it predicts, but also where the prediction is unreliable.

Uncertainty estimation is a natural reliability layer, but common methods trade accuracy, cost, and deployability. Deep ensembles, MC dropout, and test-time augmentation can be strong but require repeated inference or multiple models \cite{Lakshminarayanan2017DeepEnsembles,Gal2016DropoutBayes,Wang2019TTA}. ConfidNet instead learns to predict the true-class probability (TCP) as a surrogate for confidence, providing a lightweight learned confidence estimate but without optimizing uncertainty ranking or selective error detection \cite{NIPS2019_8556}. Temperature scaling is efficient but mainly rescales confidence \cite{Guo2017Calibration}. Deterministic methods such as DUQ, DUE, and DDU-Seg avoid repeated backbone passes, yet depend on feature geometry, density models, or training-time constraints that may be awkward for an already validated medical segmentor \cite{VanAmersfoort2020DUQ,Liu2020DUE,Mukhoti2021DDUSeg,Mukhoti2023DDU}.

This deployment setting motivates \method, a supervised uncertainty head attached to a frozen pretrained backbone. The operational premise is that unreliable delineations should be unstable under small structured perturbations of tapped features. \method{} learns a compact probe space, evaluates fixed deterministic signed perturbation patterns, and separates two uses of uncertainty: a positive calibration map that tempers logits for proper scoring rules and an unconstrained ranking map that orders voxels for selective prediction and review. We explicitly define the probe patterns and use ``single backbone pass'' to mean one expensive frozen-backbone pass plus lightweight head computations.

Our contributions are a frozen-backbone uncertainty module that preserves argmax segmentations, a deterministic perturbation-probe formulation with separate calibration and ranking maps, voxel-level evaluation on three datasets, and ablations showing the importance of the two-map design, ranking losses, and learned probes.

\section{Related Work}
\label{sec:related_work}
\paragraph{Multi-pass uncertainty.}
Deep ensembles expose disagreement across independently trained models \cite{Lakshminarayanan2017DeepEnsembles}; MC dropout retains dropout at test time \cite{Gal2016DropoutBayes}; and test-time augmentation measures variability under plausible input transforms \cite{Wang2019TTA}. They are useful references, but their cost scales with models, stochastic samples, or augmentations, which is especially relevant for volumetric images.

\paragraph{Calibration and deterministic single-pass methods.}
Temperature scaling learns a global scalar temperature on validation data and can improve probability scores, but it does not optimize spatial error ranking \cite{Guo2017Calibration}. ConfidNet learns a confidence predictor from intermediate features to regress the true-class probability (TCP), providing learned failure confidence without repeated stochastic inference \cite{NIPS2019_8556}. DUQ and DUE use distance-aware deterministic representations \cite{VanAmersfoort2020DUQ,Liu2020DUE}, while DDU and DDU-Seg fit class-conditional feature densities \cite{Mukhoti2021DDUSeg,Mukhoti2023DDU}. These methods show that single-pass uncertainty or confidence estimation is possible, but rely on learned confidence regression, representation geometry, or high-dimensional density estimation.

\paragraph{Positioning of \method.}
Latent perturbation has precedents in adversarial robustness, sensitivity analysis, and test-time perturbation methods \cite{Goodfellow2015Explaining,Wang2019TTA}. \method{} learns feature-tapped probe responses on a frozen backbone and uses finite deterministic perturbations to train two maps with different objectives, avoiding learned regression to a single confidence target, new constraints on global representation geometry, or class-conditional density fitting.

\section{Method}
\label{sec:method}

\begin{figure}[t]
    \centering
    \includegraphics[width=\textwidth]{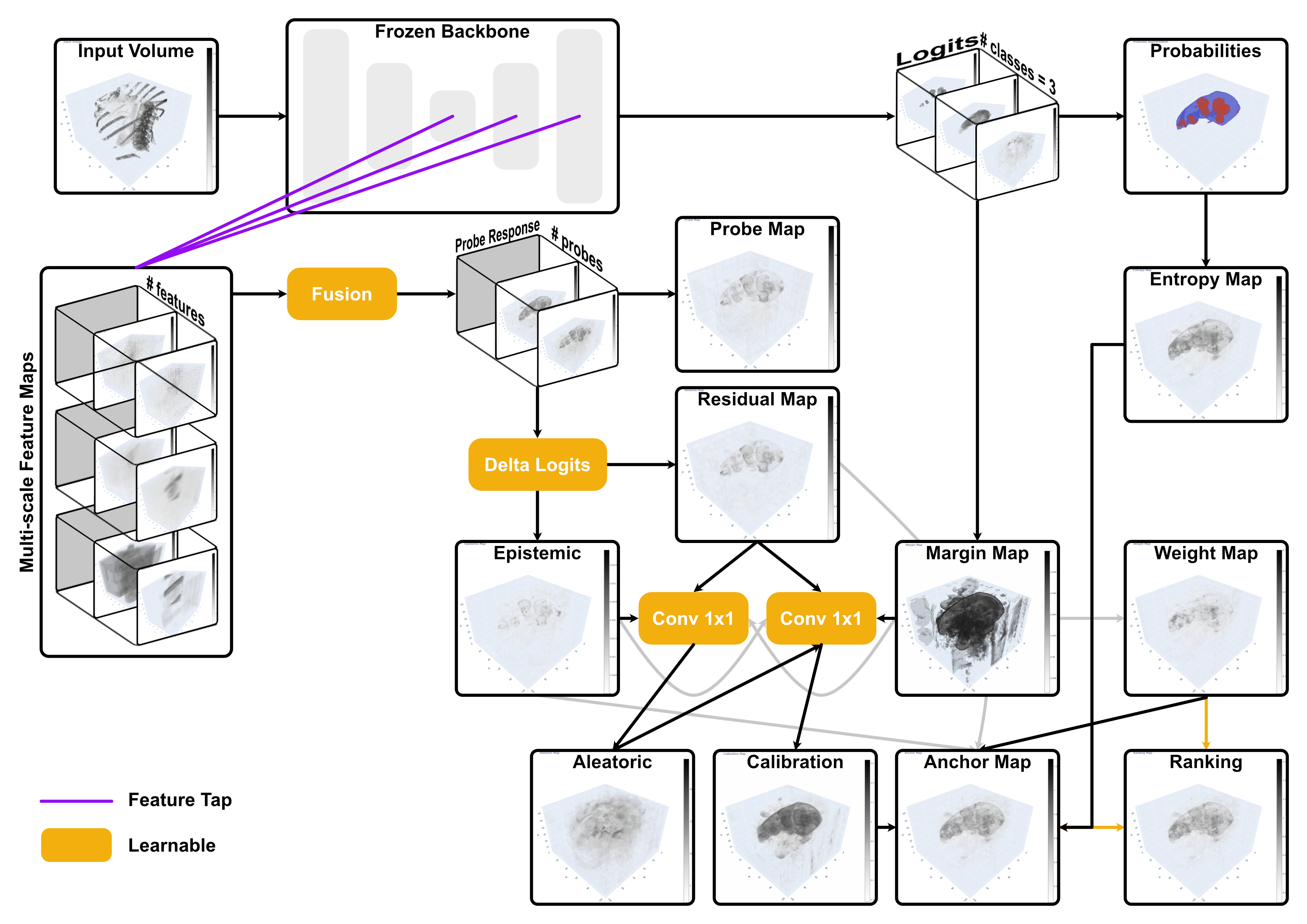}
    \caption{\textbf{Overview of \method.} A frozen segmentation backbone produces raw logits $z$ and tapped decoder features $h$. A lightweight uncertainty head maps $h$ to compact probe responses, evaluates deterministic signed perturbation patterns, and derives intermediate maps $U_{\mathrm{epi}}$, $U_{\mathrm{probe}}$, $U_{\mathrm{res}}$, and $U_{\mathrm{ale}}$. These are combined with margin and entropy cues to produce $U_{\mathrm{cal}}$ for probability tempering and $U_{\mathrm{rnk}}$ for voxel-wise error ranking.}
    \label{fig:architecture}
\end{figure}

\subsection{Compact Formulation}

Let $z=f_{\theta}(x)\in\Rset^{B\times C\times\Omega}$ be frozen-backbone logits and let $h$ be tapped decoder features after projection, resizing, concatenation, and fusion. In our DynUNet implementation, the taps come from two decoder upsampling blocks and the output block, giving both coarse semantic and finer boundary cues. \method{} leaves $z$ and the argmax mask unchanged. From $h$, a $1{\times}1$ head computes probes $v=\psi(h)\in\Rset^{B\times R\times\Omega}$ with learned scales $\sigma_r=\mathrm{softplus}(\alpha_r)+\epsilon$. Fixed signs then induce class-logit perturbations
\begin{equation}
    \Delta z(u^{(k)})=A((\sigma\odot u^{(k)})\odot v),\quad
    u^{(k)}_r=\mathrm{sign}\!\left\{\cos\!\left(\frac{\pi(k+1)(r+0.5)}{R}\right)\right\},
    \label{eq:patterns}
\end{equation}
with $R=K=8$ by default. These fixed cosine-sign directions spread probes across the learned compact basis and measure local sensitivity. The perturbed probabilities $p^{(k)}=\mathrm{softmax}(z+\Delta z(u^{(k)}))$ define
\begin{equation}
    U_{\mathrm{epi}}=\sum_{c=1}^{C}\mathrm{Var}_{k}(p^{(k)}_c),\quad
    U_{\mathrm{probe}}=\frac{1}{R}\sum_r v_r^2,\quad
    U_{\mathrm{res}}=\frac{1}{C}\sum_c \Delta z_c^2 .
\end{equation}
Single-backbone-pass inference means one expensive 3D backbone evaluation; the $K$ perturbation evaluations are lightweight uncertainty-head operations on cached features. The full formula reference is provided in the supplementary appendix.

\subsection{Calibration, Ranking, and Training}

Let $p=\mathrm{softmax}(z)$, margin $m=p_{(1)}-p_{(2)}$, ambiguity weight $w=\exp(-\gamma m)$ with default $\gamma=4$, and $U_{\mathrm{ale}}=\mathrm{softplus}(\psi_{\mathrm{ale}}(h))$. The positive calibration map is
\begin{equation}
    U_{\mathrm{cal}}=\mathrm{softplus}\{\psi_{\mathrm{cal}}[\log(1+U_{\mathrm{epi}}+U_{\mathrm{res}}),\log(1+U_{\mathrm{ale}}),m]\},
    \qquad \tilde{z}=z/\sqrt{1+U_{\mathrm{cal}}},
\end{equation}
so calibrated probabilities change but raw hard segmentations do not. Error localization uses an unconstrained score
\begin{equation}
    U_{\mathrm{rnk}}=(1+0.1\tanh(a))U_{\mathrm{anchor}}+b+\mathrm{softplus}(c)w,
\end{equation}
where $U_{\mathrm{anchor}}=\log(1+U_{\mathrm{epi}})+0.5\log(1+U_{\mathrm{res}})+0.25\log(1+U_{\mathrm{cal}})+0.25H(p)/\log C+w$. Thus $U_{\mathrm{cal}}$ supports probabilistic scoring and $U_{\mathrm{rnk}}$ ranks likely segmentation errors.

The segmentation backbone is frozen and $\lambda_{\mathrm{seg}}=0$, so the uncertainty head cannot improve Dice by modifying masks. We optimize only the uncertainty head with
\begin{equation}
    \mathcal{L}=0.5\mathcal{L}_{\mathrm{NLL}}+0.25(\mathcal{L}_{\mathrm{EC}}+\mathcal{L}_{\mathrm{Pairwise}}+\mathcal{L}_{\mathrm{Tail}})+0.05(\mathcal{L}_{\mathrm{Trust}}+\mathcal{L}_{\mathrm{Anchor}}+\mathcal{L}_{\mathrm{Residual}}).
    \label{eq:objective}
\end{equation}
$\mathcal{L}_{\mathrm{NLL}}$ uses $\tilde{z}$. Ranking losses use $u_i=U_{\mathrm{rnk},i}$ and error labels $e_i=\indicator[\hat{y}_i\ne y_i]$, with $\hat{y}_i=\arg\max_c z_{i,c}$. These include error-correlation BCE on standardized scores, an erroneous/correct pairwise margin, and a tail penalty for erroneous voxels assigned low uncertainty. Trust, anchor, and residual terms regularize perturbation magnitude, agreement with the anchor, and residual energy on easy voxels.

\section{Experiments}
\label{sec:experiments}

\subsection{Datasets, Backbones, and Baselines}

We evaluate on ACDC cardiac MRI \cite{Bernard2018DeepLearningCardiacMRI}, BraTS2024 multimodal brain MRI \cite{deverdier20242024braintumorsegmentation}, and LiTS abdominal CT \cite{Bilic2019LiTS}. Because BraTS2024 has no annotated public test set, we sample 200 training and 100 held-out test cases from the public training set; for LiTS, we reserve the final 10 cases for testing. These splits are not site-, scanner-, or vendor-held-out splits.

All segmentation backbones are 3D DynUNet models trained on five non-overlapping folds plus one ``all-data'' model using MONAI tooling \cite{Cardoso2022MONAI}. The five-fold Dice ranges were $88.21$--$90.51\%$ on ACDC, $55.87$--$60.07\%$ on BraTS2024, and $71.77$--$76.75\%$ on LiTS. The corresponding all-data Dice scores were $90.35\%$, $62.75\%$, and $78.21\%$. The BraTS Dice range frames the uncertainty comparison under this fixed-backbone protocol and should be considered when interpreting absolute mask quality.

ACDC is resampled to a $64{\times}128{\times}128$ patch because of coarse through-plane spacing; BraTS2024 and LiTS use $128^3$ patches. All backbones use Dice+CE training with 3D sliding-window inference and Gaussian blending. \method{} trains only the second-stage uncertainty head with AdamW, cosine learning-rate scheduling, gradient clipping, early stopping, and the default $R=K=8$ probe configuration.

We compare \method{} to Deep Ensembles, TTA, MC Dropout, TS, DUQ, DDU-Seg, DUE, and ConfidNet. Ranking uses each method's native scalar uncertainty: ensemble mutual information; predictive-distribution uncertainty for MC Dropout and TTA; entropy after TS; native scores for DUQ, DDU-Seg, and DUE; and $U_{\mathrm{rnk}}$ for \method. For probabilistic scores, \method{} uses probabilities tempered by $U_{\mathrm{cal}}$; AUROC and AURC use $U_{\mathrm{rnk}}$ and raw-logit hard errors.

\paragraph{Calibration data.}
\method{} trains a supervised second-stage uncertainty head on top of the frozen backbone. Our protocol uses the train/test separation above; in deployment, labeled validation data should be reserved for the uncertainty head and temperature-style baselines.

\subsection{Voxel-Level Metrics}

Metrics are voxel-level over the test set. Let $\hat{y}_i=\arg\max_c z_{i,c}$ and $e_i=\indicator[\hat{y}_i\ne y_i]$. Dice reports overlap; multi-class Brier is mean squared error between probabilities and one-hot labels; and AUROC tests whether uncertainty $u_i$ ranks erroneous voxels above correct voxels. For AURC, voxels are sorted from low to high uncertainty, the most confident $\alpha$ fraction is retained, and residual risk is
\begin{equation}
    \mathrm{Risk}(\alpha)=
    \frac{1}{\lfloor \alpha N \rfloor}
    \sum_{j=1}^{\lfloor \alpha N \rfloor} e_{\pi(j)} .
\end{equation}
AURC is the numerical area under this risk--coverage curve. Lower AURC is better, and all table values are reported in $10^{-4}$ units.

\subsection{Main Results}

\begin{table}[t]
\centering
\scriptsize
\setlength{\tabcolsep}{4pt}
\begin{tabular}{lcccc}
\toprule
\textbf{Method} & \textbf{Dice} $\uparrow$ & \textbf{Brier} $\downarrow$ & \textbf{AUROC} $\uparrow$ & \textbf{AURC} $\downarrow$ \\
\midrule
\multicolumn{5}{c}{\textbf{ACDC}}\\
\midrule
Deep Ensembles & $\mathbf{0.9061{\pm}0.0040}$ & $\mathbf{0.0107{\pm}0.0005}$ & $0.9813{\pm}0.0017$ & $2.5326{\pm}0.5455$ \\
TTA & $0.6607{\pm}0.0130$ & $0.0251{\pm}0.0013$ & $0.9826{\pm}0.0016$ & $8.1500{\pm}1.2456$ \\
MC Dropout & $0.9010{\pm}0.0048$ & $0.0123{\pm}0.0007$ & $0.9549{\pm}0.0040$ & $6.3111{\pm}1.2639$ \\
TS & $0.9035{\pm}0.0044$ & $0.0115{\pm}0.0006$ & $0.9714{\pm}0.0034$ & $4.6798{\pm}1.0839$ \\
DUQ & $0.8871{\pm}0.0064$ & $0.0141{\pm}0.0013$ & $0.9735{\pm}0.0023$ & $4.7843{\pm}0.9290$ \\
DDU-Seg & $0.9035{\pm}0.0044$ & $0.0122{\pm}0.0006$ & $0.9716{\pm}0.0034$ & $4.6628{\pm}1.0842$ \\
DUE & $0.9024{\pm}0.0044$ & $0.0115{\pm}0.0008$ & $0.9809{\pm}0.0020$ & $3.0233{\pm}0.5694$ \\
ConfidNet & $0.9035{\pm}0.0044$ & $0.0122{\pm}0.0006$ & $0.9778{\pm}0.0028$ & $3.3990{\pm}0.7888$ \\
\textbf{\method} & $0.9035{\pm}0.0044$ & $0.0113{\pm}0.0006$ & $\mathbf{0.9838{\pm}0.0022}$ & $\mathbf{2.4885{\pm}0.6077}$ \\
\midrule
\multicolumn{5}{c}{\textbf{BraTS2024}}\\
\midrule
Deep Ensembles & $0.5988{\pm}0.0133$ & $0.0042{\pm}0.0004$ & $0.9917{\pm}0.0011$ & $0.3898{\pm}0.0697$ \\
TTA & $0.6131{\pm}0.0138$ & $0.0041{\pm}0.0003$ & $0.9919{\pm}0.0007$ & $0.3106{\pm}0.0372$ \\
MC Dropout & $0.5759{\pm}0.0153$ & $0.0064{\pm}0.0008$ & $0.9672{\pm}0.0028$ & $1.9627{\pm}0.3476$ \\
TS & $\mathbf{0.6275{\pm}0.0130}$ & $0.0038{\pm}0.0003$ & $0.9846{\pm}0.0017$ & $0.5293{\pm}0.0786$ \\
DUQ & $0.6031{\pm}0.0112$ & $\mathbf{0.0036{\pm}0.0003}$ & $0.9912{\pm}0.0018$ & $0.4836{\pm}0.1587$ \\
DDU-Seg & $\mathbf{0.6275{\pm}0.0130}$ & $0.0040{\pm}0.0003$ & $0.9805{\pm}0.0026$ & $0.5009{\pm}0.0701$ \\
DUE & $0.5864{\pm}0.0089$ & $\mathbf{0.0036{\pm}0.0003}$ & $0.9932{\pm}0.0008$ & $0.2951{\pm}0.0552$ \\
ConfidNet & $\mathbf{0.6275{\pm}0.0131}$ & $0.0040{\pm}0.0003$ & $0.9901{\pm}0.0015$ & $0.5178{\pm}0.1231$ 
\\
\textbf{\method} & $\mathbf{0.6275{\pm}0.0130}$ & $0.0037{\pm}0.0003$ & $\mathbf{0.9946{\pm}0.0007}$ & $\mathbf{0.2660{\pm}0.0528}$ \\
\midrule
\multicolumn{5}{c}{\textbf{LiTS}}\\
\midrule
Deep Ensembles & $0.7932{\pm}0.0336$ & $0.0063{\pm}0.0022$ & $0.9920{\pm}0.0027$ & $0.9628{\pm}0.6615$ \\
TTA & $0.7702{\pm}0.0362$ & $0.0077{\pm}0.0020$ & $0.9895{\pm}0.0030$ & $1.1241{\pm}0.6090$ \\
MC Dropout & $0.7823{\pm}0.0342$ & $0.0085{\pm}0.0023$ & $0.9520{\pm}0.0082$ & $3.4640{\pm}1.3448$ \\
TS & $0.7821{\pm}0.0343$ & $0.0068{\pm}0.0020$ & $0.9861{\pm}0.0042$ & $1.6388{\pm}1.0165$ \\
DUQ & $0.7629{\pm}0.0289$ & $0.D0065{\pm}0.0023$ & $0.9915{\pm}0.0029$ & $0.9633{\pm}0.6536$ \\
DDU-Seg & $0.7821{\pm}0.0343$ & $0.0069{\pm}0.0021$ & $0.9857{\pm}0.0043$ & $1.6514{\pm}1.0159$ \\
DUE & $\mathbf{0.8016{\pm}0.0350}$ & $\mathbf{0.0057{\pm}0.0021}$ & $0.9898{\pm}0.0049$ & $1.3972{\pm}1.0533$ \\
ConfidNet & $0.7841{\pm}0.0361$ & $0.0069{\pm}0.0022$ & $0.9878{\pm}0.0048$ & $1.4832{\pm}1.0382$ \\
\textbf{\method} & $0.7821{\pm}0.0343$ & $0.0067{\pm}0.0020$ & $\mathbf{0.9925{\pm}0.0025}$ & $\mathbf{0.8193{\pm}0.5117}$ \\
\bottomrule
\end{tabular}
\caption{\textbf{Main quantitative results.} Mean $\pm$ standard deviation over test cases. AURC is reported in $10^{-4}$ units. Bold indicates the best value in each dataset/metric. Multi-pass methods are high-cost references; the main comparison for \method{} is the single-backbone-pass group.}
\label{tab:main_results}
\end{table}

Table~\ref{tab:main_results} shows that \method{} is the most consistent single-backbone-pass method for ranking uncertainty. It achieves the best AUROC and AURC on ACDC, BraTS2024, and LiTS while preserving the frozen-backbone Dice. On ACDC, \method{} improves AURC over DUE by $17.7\%$ ($3.0233$ to $2.4885$) and slightly over Deep Ensembles ($2.5326$ to $2.4885$), despite using one backbone pass rather than five models. On BraTS2024, it improves AURC over DUE by $9.9\%$ and over TTA by $14.4\%$, while matching the all-data backbone Dice. On LiTS, it gives the lowest AURC ($0.8193$) and highest AUROC ($0.9925$), although DUE has the best Dice and Brier. The main gain is stronger voxel ordering for selective prediction, while segmentation quality and calibration remain tied to the frozen backbone and calibration map. Multi-pass baselines remain important references and are sometimes competitive, but require repeated inference; on BraTS2024, Deep Ensembles also inherit weaker fold-specific backbones.

\subsection{Qualitative Analysis}

\begin{figure}[t]
\centering
\setlength{\tabcolsep}{2pt}
\begin{tabular}{ccc}
\includegraphics[width=.32\textwidth]{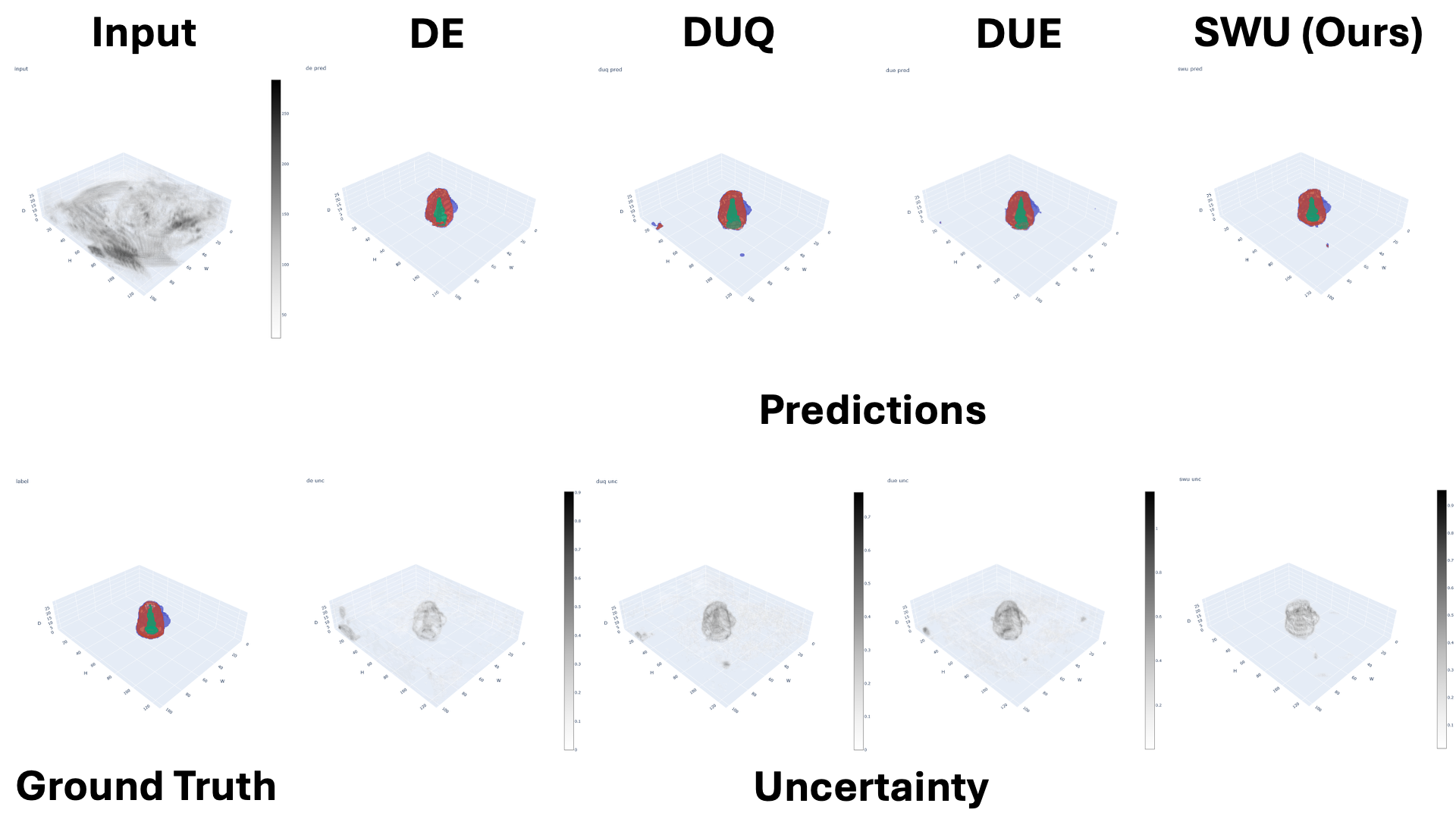} &
\includegraphics[width=.32\textwidth]{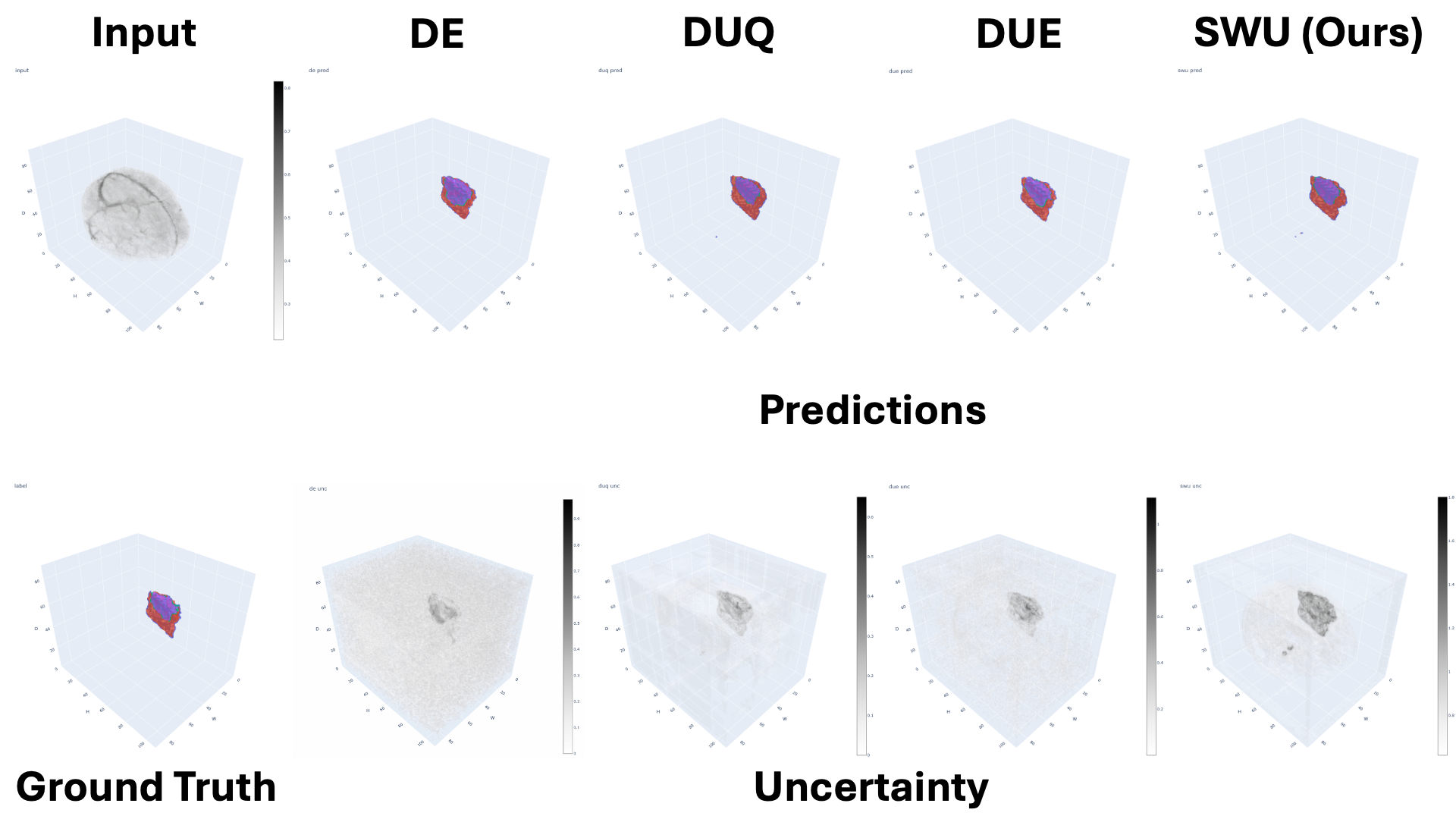} &
\includegraphics[width=.32\textwidth]{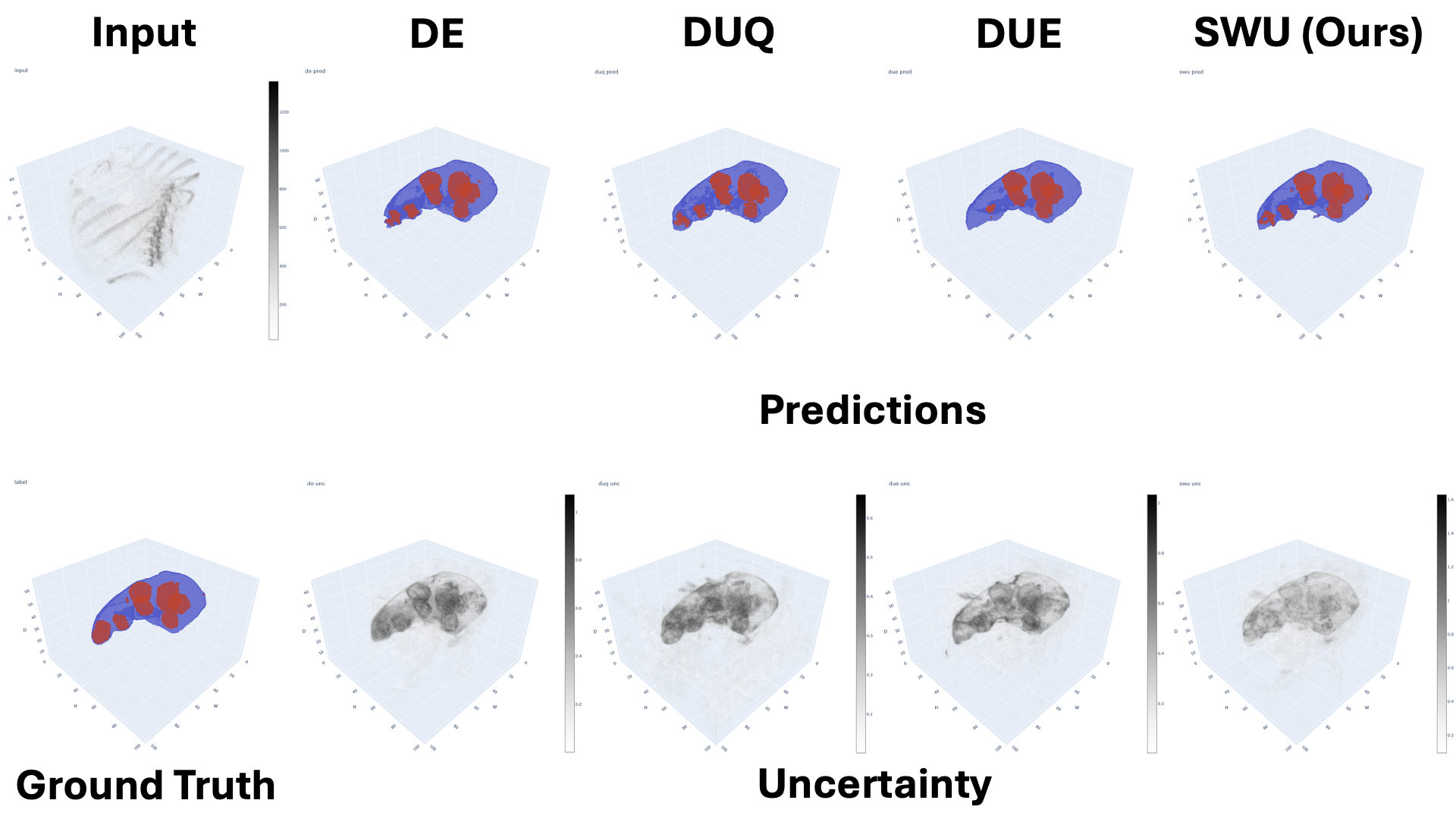} \\
{\scriptsize ACDC case 7} & {\scriptsize BraTS2024 case 44} & {\scriptsize LiTS case 8}
\end{tabular}
\caption{\textbf{Representative qualitative cases.} Each panel shows the input, predicted masks and uncertainty maps for Deep Ensembles, DUQ, DUE, and \method, together with the ground truth. \method{} concentrates uncertainty near predicted anatomy and reduces isolated high-uncertainty regions away from anatomy on these cases.}
\label{fig:qualitative_cases}
\end{figure}

Figure~\ref{fig:qualitative_cases} shows that \method{} tends to attach high uncertainty to anatomical boundaries and failure regions rather than diffuse background. Table~\ref{tab:qualitative_cases} gives the corresponding case metrics for the displayed DE/DUQ/DUE/\method{} comparison. On ACDC case 7, Deep Ensembles produce a broad halo and DUQ introduces isolated high-uncertainty components away from the contour, whereas \method{} keeps uncertainty closer to the cardiac contour and is best on all four metrics. On BraTS2024 case 44, the lesion is irregular and all methods have limited Dice, but \method{} places uncertainty near the lesion boundary, gives the best Dice, and remains close to DUQ/DUE on ranking. On LiTS case 8, all methods show liver-boundary uncertainty; \method{} produces fewer isolated high-uncertainty regions away from the liver and lesion, emphasizes the lesion region, and is best on Dice/AUROC/AURC while remaining close to the best Brier value.

\begin{table}[t]
\centering
\scriptsize
\setlength{\tabcolsep}{2.4pt}
\begin{tabular}{lccc}
\toprule
\textbf{Method} & \textbf{ACDC 7} & \textbf{BraTS2024 44} & \textbf{LiTS 8}\\
\midrule
DE & $.8755/.0093/.9842/1.2000$ & $.5201/.0020/.9921/.1132$ & $.9218/.0231/.9690/7.1483$\\
DUQ & $.8450/.0117/.9848/1.6000$ & $.5530/.0015/.9982/.0228$ & $.8937/.0251/.9673/7.0911$\\
DUE & $.8696/.0098/.9838/1.4100$ & $.5550/.0015/.9984/.0179$ & $.9221/.0219/.9441/11.2840$\\
\textbf{\method} & $\mathbf{.8867/.0087/.9905/.7100}$ & $\mathbf{.5554}/.0017/.9974/.0347$ & $\mathbf{.9254}/.0222/\mathbf{.9695}/\mathbf{5.5866}$\\
\bottomrule
\end{tabular}
\caption{\textbf{Selected-case metrics.} Cells report Dice/Brier/AUROC/AURC; AURC is in $10^{-4}$.}
\label{tab:qualitative_cases}
\end{table}

\subsection{Statistical Testing and Significance Scoring}

Case-wise paired two-sided Wilcoxon tests compare methods within each dataset--metric family. $p$-values are Holm-corrected over all pairwise method tests; $\alpha=0.05$. Directions follow the metric: higher Dice/AUROC and lower Brier/AURC are better. For method $m$ and dataset $d$, \(S_d(m)=\sum_{q,o}(\indicator[m\succ_{d,q}o]-\indicator[o\succ_{d,q}m])\), where $q$ indexes metrics, $o$ competitors, and $\succ$ a corrected significant win. Scores for (DE, TTA, MCDO, TS, DUQ, DDU-Seg, DUE, ConfidNet, \method) are ACDC (+16,-20,-20,-1,-15,-5,+14,+6,+25) and BraTS2024 (-2,+3,-29,-5,+6,-6,+9,0,+24). LiTS has $n=10$ and is set to $0$ for underpowered scoring. Aggregate scores are (+14,-17,-49,-6,-9,-11,+23,+6,+49); \method{} is never significantly outperformed on ACDC/BraTS2024. Tables 4, 5, and 6 contain exact $p$-values for each dataset.

\begin{table*}[!t]
\centering
\let\textbf\textnormal
\fontsize{5.5}{5.7}\selectfont
\setlength{\tabcolsep}{1.3pt}
\renewcommand{\arraystretch}{0.44}
\begin{tabular*}{0.92\textwidth}{@{\extracolsep{\fill}}l|rrrrrrrrr@{}}
\toprule
\textbf{Method} & \textbf{DE} & \textbf{TTA} & \textbf{MCDO} & \textbf{TS} & \textbf{DUQ} & \textbf{DDU-Seg} & \textbf{DUE} & \textbf{ConfidNet} & \textbf{SWU} \\
\midrule
\multicolumn{10}{c}{\textbf{Dice} $\uparrow$} \\
\midrule
\textbf{DE} & -- & \verySmallP{} & 0.0104 & 1.0000 & \verySmallP{} & 1.0000 & 1.0000 & 1.0000 & 1.0000 \\
\textbf{TTA} & \verySmallP{} & -- & \verySmallP{} & \verySmallP{} & \verySmallP{} & \verySmallP{} & \verySmallP{} & \verySmallP{} & \verySmallP{} \\
\textbf{MCDO} & 0.0104 & \verySmallP{} & -- & 0.0004 & \verySmallP{} & 0.0004 & 1.0000 & 0.0007 & 0.0004 \\
\textbf{TS} & 1.0000 & \verySmallP{} & 0.0004 & -- & \verySmallP{} & 1.0000 & 1.0000 & 1.0000 & 1.0000 \\
\textbf{DUQ} & \verySmallP{} & \verySmallP{} & \verySmallP{} & \verySmallP{} & -- & \verySmallP{} & \verySmallP{} & \verySmallP{} & \verySmallP{} \\
\textbf{DDU-Seg} & 1.0000 & \verySmallP{} & 0.0004 & 1.0000 & \verySmallP{} & -- & 1.0000 & 1.0000 & 1.0000 \\
\textbf{DUE} & 1.0000 & \verySmallP{} & 1.0000 & 1.0000 & \verySmallP{} & 1.0000 & -- & 1.0000 & 1.0000 \\
\textbf{ConfidNet} & 1.0000 & \verySmallP{} & 0.0007 & 1.0000 & \verySmallP{} & 1.0000 & 1.0000 & -- & 1.0000 \\
\textbf{SWU} & 1.0000 & \verySmallP{} & 0.0004 & 1.0000 & \verySmallP{} & 1.0000 & 1.0000 & 1.0000 & -- \\
\midrule
\multicolumn{10}{c}{\textbf{Brier} $\downarrow$} \\
\midrule
\textbf{DE} & -- & \verySmallP{} & \verySmallP{} & 0.0017 & \verySmallP{} & \verySmallP{} & 1.0000 & \verySmallP{} & 0.1152 \\
\textbf{TTA} & \verySmallP{} & -- & \verySmallP{} & \verySmallP{} & \verySmallP{} & \verySmallP{} & \verySmallP{} & \verySmallP{} & \verySmallP{} \\
\textbf{MCDO} & \verySmallP{} & \verySmallP{} & -- & \verySmallP{} & 1.0000 & 1.0000 & \verySmallP{} & 1.0000 & \verySmallP{} \\
\textbf{TS} & 0.0017 & \verySmallP{} & \verySmallP{} & -- & 0.0073 & \verySmallP{} & 0.1085 & \verySmallP{} & \verySmallP{} \\
\textbf{DUQ} & \verySmallP{} & \verySmallP{} & 1.0000 & 0.0073 & -- & 1.0000 & \verySmallP{} & 1.0000 & 0.0006 \\
\textbf{DDU-Seg} & \verySmallP{} & \verySmallP{} & 1.0000 & \verySmallP{} & 1.0000 & -- & \verySmallP{} & 1.0000 & \verySmallP{} \\
\textbf{DUE} & 1.0000 & \verySmallP{} & \verySmallP{} & 0.1085 & \verySmallP{} & \verySmallP{} & -- & \verySmallP{} & 0.5805 \\
\textbf{ConfidNet} & \verySmallP{} & \verySmallP{} & 1.0000 & \verySmallP{} & 1.0000 & 1.0000 & \verySmallP{} & -- & \verySmallP{} \\
\textbf{SWU} & 0.1152 & \verySmallP{} & \verySmallP{} & \verySmallP{} & 0.0006 & \verySmallP{} & 0.5805 & \verySmallP{} & -- \\

\midrule
\multicolumn{10}{c}{\textbf{AUROC} $\uparrow$} \\
\midrule
\textbf{DE} & -- & 0.6300 & \verySmallP{} & \verySmallP{} & 0.0002 & \verySmallP{} & 1.0000 & 1.0000 & \verySmallP{} \\
\textbf{TTA} & 0.6300 & -- & \verySmallP{} & \verySmallP{} & \verySmallP{} & \verySmallP{} & 0.8181 & 0.0407 & 0.0021 \\
\textbf{MCDO} & \verySmallP{} & \verySmallP{} & -- & \verySmallP{} & \verySmallP{} & \verySmallP{} & \verySmallP{} & \verySmallP{} & \verySmallP{} \\
\textbf{TS} & \verySmallP{} & \verySmallP{} & \verySmallP{} & -- & 1.0000 & 0.0335 & \verySmallP{} & \verySmallP{} & \verySmallP{} \\
\textbf{DUQ} & 0.0002 & \verySmallP{} & \verySmallP{} & 1.0000 & -- & 1.0000 & \verySmallP{} & 0.0012 & \verySmallP{} \\
\textbf{DDU-Seg} & \verySmallP{} & \verySmallP{} & \verySmallP{} & 0.0335 & 1.0000 & -- & \verySmallP{} & \verySmallP{} & \verySmallP{} \\
\textbf{DUE} & 1.0000 & 0.8181 & \verySmallP{} & \verySmallP{} & \verySmallP{} & \verySmallP{} & -- & 1.0000 & \verySmallP{} \\
\textbf{ConfidNet} & 1.0000 & 0.0407 & \verySmallP{} & \verySmallP{} & 0.0012 & \verySmallP{} & 1.0000 & -- & \verySmallP{} \\
\textbf{SWU} & \verySmallP{} & 0.0021 & \verySmallP{} & \verySmallP{} & \verySmallP{} & \verySmallP{} & \verySmallP{} & \verySmallP{} & -- \\

\midrule
\multicolumn{10}{c}{\textbf{AURC} $\downarrow$} \\
\midrule
\textbf{DE} & -- & \verySmallP{} & \verySmallP{} & \verySmallP{} & \verySmallP{} & \verySmallP{} & 0.9890 & 0.2670 & \verySmallP{} \\
\textbf{TTA} & \verySmallP{} & -- & \verySmallP{} & \verySmallP{} & \verySmallP{} & \verySmallP{} & \verySmallP{} & \verySmallP{} & \verySmallP{} \\
\textbf{MCDO} & \verySmallP{} & \verySmallP{} & -- & \verySmallP{} & 0.0010 & \verySmallP{} & \verySmallP{} & \verySmallP{} & \verySmallP{} \\
\textbf{TS} & \verySmallP{} & \verySmallP{} & \verySmallP{} & -- & 0.7355 & 0.0195 & \verySmallP{} & \verySmallP{} & \verySmallP{} \\
\textbf{DUQ} & \verySmallP{} & \verySmallP{} & 0.0010 & 0.7355 & -- & 0.7355 & \verySmallP{} & 0.0006 & \verySmallP{} \\
\textbf{DDU-Seg} & \verySmallP{} & \verySmallP{} & \verySmallP{} & 0.0195 & 0.7355 & -- & 0.0001 & \verySmallP{} & \verySmallP{} \\
\textbf{DUE} & 0.9890 & \verySmallP{} & \verySmallP{} & \verySmallP{} & \verySmallP{} & 0.0001 & -- & 1.0000 & 0.0012 \\
\textbf{ConfidNet} & 0.2670 & \verySmallP{} & \verySmallP{} & \verySmallP{} & 0.0006 & \verySmallP{} & 1.0000 & -- & \verySmallP{} \\
\textbf{SWU} & \verySmallP{} & \verySmallP{} & \verySmallP{} & \verySmallP{} & \verySmallP{} & \verySmallP{} & 0.0012 & \verySmallP{} & -- \\

\bottomrule
\end{tabular*}
\caption{Holm-corrected pairwise Wilcoxon $p$-values for ACDC across all metrics. Off-diagonal entries compare row and column methods across test cases; the matrix is symmetric, and \verySmallP{} denotes $p<0.0001$.}
\label{tab:pvalues_acdc}
\end{table*}
\vspace{0pt}
\begin{table*}[!t]
\centering
\let\textbf\textnormal
\fontsize{5.3}{5.5}\selectfont
\setlength{\tabcolsep}{1.3pt}
\renewcommand{\arraystretch}{0.40}
\begin{tabular*}{0.92\textwidth}{@{\extracolsep{\fill}}l|rrrrrrrrr@{}}
\toprule
\textbf{Method} & \textbf{DE} & \textbf{TTA} & \textbf{MCDO} & \textbf{TS} & \textbf{DUQ} & \textbf{DDU-Seg} & \textbf{DUE} & \textbf{ConfidNet} & \textbf{SWU} \\
\midrule
\multicolumn{10}{c}{\textbf{Dice} $\uparrow$} \\
\midrule
\textbf{DE} & -- & 0.3782 & 0.3541 & \verySmallP{} & 1.0000 & \verySmallP{} & 1.0000 & 0.0001 & \verySmallP{} \\
\textbf{TTA} & 0.3782 & -- & 0.0230 & 0.1401 & 0.6299 & 0.1401 & 0.2435 & 0.2241 & 0.1401 \\
\textbf{MCDO} & 0.3541 & 0.0230 & -- & \verySmallP{} & 0.6536 & \verySmallP{} & 1.0000 & \verySmallP{} & \verySmallP{} \\
\textbf{TS} & \verySmallP{} & 0.1401 & \verySmallP{} & -- & \verySmallP{} & 1.0000 & 0.0002 & 1.0000 & 1.0000 \\
\textbf{DUQ} & 1.0000 & 0.6299 & 0.6536 & \verySmallP{} & -- & \verySmallP{} & 0.6299 & \verySmallP{} & \verySmallP{} \\
\textbf{DDU-Seg} & \verySmallP{} & 0.1401 & \verySmallP{} & 1.0000 & \verySmallP{} & -- & 0.0002 & 1.0000 & 1.0000 \\
\textbf{DUE} & 1.0000 & 0.2435 & 1.0000 & 0.0002 & 0.6299 & 0.0002 & -- & 0.0003 & 0.0002 \\
\textbf{ConfidNet} & 0.0001 & 0.2241 & \verySmallP{} & 1.0000 & \verySmallP{} & 1.0000 & 0.0003 & -- & 1.0000 \\
\textbf{SWU} & \verySmallP{} & 0.1401 & \verySmallP{} & 1.0000 & \verySmallP{} & 1.0000 & 0.0002 & 1.0000 & -- \\

\midrule
\multicolumn{10}{c}{\textbf{Brier} $\downarrow$} \\
\midrule
\textbf{DE} & -- & 1.0000 & \verySmallP{} & 0.2642 & 0.0020 & 1.0000 & 0.0151 & 1.0000 & 0.0141 \\
\textbf{TTA} & 1.0000 & -- & 0.0003 & 0.1761 & 0.0083 & 1.0000 & 0.0032 & 1.0000 & 0.0010 \\
\textbf{MCDO} & \verySmallP{} & 0.0003 & -- & \verySmallP{} & \verySmallP{} & \verySmallP{} & \verySmallP{} & \verySmallP{} & \verySmallP{} \\
\textbf{TS} & 0.2642 & 0.1761 & \verySmallP{} & -- & 0.0151 & \verySmallP{} & 0.2642 & \verySmallP{} & \verySmallP{} \\
\textbf{DUQ} & 0.0020 & 0.0083 & \verySmallP{} & 0.0151 & -- & \verySmallP{} & 1.0000 & \verySmallP{} & 1.0000 \\
\textbf{DDU-Seg} & 1.0000 & 1.0000 & \verySmallP{} & \verySmallP{} & \verySmallP{} & -- & 0.0020 & 1.0000 & \verySmallP{} \\
\textbf{DUE} & 0.0151 & 0.0032 & \verySmallP{} & 0.2642 & 1.0000 & 0.0020 & -- & 0.0032 & 1.0000 \\
\textbf{ConfidNet} & 1.0000 & 1.0000 & \verySmallP{} & \verySmallP{} & \verySmallP{} & 1.0000 & 0.0032 & -- & \verySmallP{} \\
\textbf{SWU} & 0.0141 & 0.0010 & \verySmallP{} & \verySmallP{} & 1.0000 & \verySmallP{} & 1.0000 & \verySmallP{} & -- \\

\midrule
\multicolumn{10}{c}{\textbf{AUROC} $\uparrow$} \\
\midrule
\textbf{DE} & -- & 0.4765 & \verySmallP{} & \verySmallP{} & 0.1205 & \verySmallP{} & 0.2419 & 0.1171 & \verySmallP{} \\
\textbf{TTA} & 0.4765 & -- & \verySmallP{} & \verySmallP{} & 0.0644 & \verySmallP{} & 0.2419 & 1.0000 & \verySmallP{} \\
\textbf{MCDO} & \verySmallP{} & \verySmallP{} & -- & \verySmallP{} & \verySmallP{} & \verySmallP{} & \verySmallP{} & \verySmallP{} & \verySmallP{} \\
\textbf{TS} & \verySmallP{} & \verySmallP{} & \verySmallP{} & -- & \verySmallP{} & \verySmallP{} & \verySmallP{} & \verySmallP{} & \verySmallP{} \\
\textbf{DUQ} & 0.1205 & 0.0644 & \verySmallP{} & \verySmallP{} & -- & \verySmallP{} & 0.6181 & 0.0351 & 0.0006 \\
\textbf{DDU-Seg} & \verySmallP{} & \verySmallP{} & \verySmallP{} & \verySmallP{} & \verySmallP{} & -- & \verySmallP{} & \verySmallP{} & \verySmallP{} \\
\textbf{DUE} & 0.2419 & 0.2419 & \verySmallP{} & \verySmallP{} & 0.6181 & \verySmallP{} & -- & 0.0184 & 0.0016 \\
\textbf{ConfidNet} & 0.1171 & 1.0000 & \verySmallP{} & \verySmallP{} & 0.0351 & \verySmallP{} & 0.0184 & -- & \verySmallP{} \\
\textbf{SWU} & \verySmallP{} & \verySmallP{} & \verySmallP{} & \verySmallP{} & 0.0006 & \verySmallP{} & 0.0016 & \verySmallP{} & -- \\

\midrule
\multicolumn{10}{c}{\textbf{AURC} $\downarrow$} \\
\midrule
\textbf{DE} & -- & 1.0000 & \verySmallP{} & 0.0136 & 0.0002 & 0.0251 & 0.0001 & 1.0000 & \verySmallP{} \\
\textbf{TTA} & 1.0000 & -- & \verySmallP{} & \verySmallP{} & 0.0412 & \verySmallP{} & 0.1420 & 0.8451 & \verySmallP{} \\
\textbf{MCDO} & \verySmallP{} & \verySmallP{} & -- & \verySmallP{} & \verySmallP{} & \verySmallP{} & \verySmallP{} & \verySmallP{} & \verySmallP{} \\
\textbf{TS} & 0.0136 & \verySmallP{} & \verySmallP{} & -- & \verySmallP{} & 1.0000 & \verySmallP{} & \verySmallP{} & \verySmallP{} \\
\textbf{DUQ} & 0.0002 & 0.0412 & \verySmallP{} & \verySmallP{} & -- & \verySmallP{} & 1.0000 & 0.0834 & 0.2562 \\
\textbf{DDU-Seg} & 0.0251 & \verySmallP{} & \verySmallP{} & 1.0000 & \verySmallP{} & -- & \verySmallP{} & \verySmallP{} & \verySmallP{} \\
\textbf{DUE} & 0.0001 & 0.1420 & \verySmallP{} & \verySmallP{} & 1.0000 & \verySmallP{} & -- & 0.0353 & 0.2974 \\
\textbf{ConfidNet} & 1.0000 & 0.8451 & \verySmallP{} & \verySmallP{} & 0.0834 & \verySmallP{} & 0.0353 & -- & \verySmallP{} \\
\textbf{SWU} & \verySmallP{} & \verySmallP{} & \verySmallP{} & \verySmallP{} & 0.2562 & \verySmallP{} & 0.2974 & \verySmallP{} & -- \\

\bottomrule
\end{tabular*}
\caption{Holm-corrected pairwise Wilcoxon $p$-values for BraTS2024 across all metrics. Off-diagonal entries compare row and column methods across test cases; the matrix is symmetric, and \verySmallP{} denotes $p<0.0001$.}
\label{tab:pvalues_brats}
\end{table*}

\begin{table*}[!t]
\centering
\let\textbf\textnormal
\fontsize{5.3}{5.5}\selectfont
\setlength{\tabcolsep}{1.3pt}
\renewcommand{\arraystretch}{0.40}
\begin{tabular*}{0.92\textwidth}{@{\extracolsep{\fill}}l|rrrrrrrrr@{}}
\toprule
\textbf{Method} & \textbf{DE} & \textbf{TTA} & \textbf{MCDO} & \textbf{TS} & \textbf{DUQ} & \textbf{DDU-Seg} & \textbf{DUE} & \textbf{ConfidNet} & \textbf{SWU} \\
\midrule
\multicolumn{10}{c}{\textbf{Dice} $\uparrow$} \\
\midrule
\textbf{DE} & -- & 0.3691 & 1.0000 & 1.0000 & 1.0000 & 1.0000 & 1.0000 & 1.0000 & 1.0000 \\
\textbf{TTA} & 0.3691 & -- & 0.9648 & 1.0000 & 1.0000 & 1.0000 & 1.0000 & 1.0000 & 1.0000 \\
\textbf{MCDO} & 1.0000 & 0.9648 & -- & 1.0000 & 0.9648 & 1.0000 & 1.0000 & 1.0000 & 1.0000 \\
\textbf{TS} & 1.0000 & 1.0000 & 1.0000 & -- & 0.9648 & 1.0000 & 1.0000 & 1.0000 & 1.0000 \\
\textbf{DUQ} & 1.0000 & 1.0000 & 0.9648 & 0.9648 & -- & 0.9648 & 0.0547 & 0.2637 & 0.9648 \\
\textbf{DDU-Seg} & 1.0000 & 1.0000 & 1.0000 & 1.0000 & 0.9648 & -- & 1.0000 & 1.0000 & 1.0000 \\
\textbf{DUE} & 1.0000 & 1.0000 & 1.0000 & 1.0000 & 0.0547 & 1.0000 & -- & 0.8398 & 1.0000 \\
\textbf{ConfidNet} & 1.0000 & 1.0000 & 1.0000 & 1.0000 & 0.2637 & 1.0000 & 0.8398 & -- & 1.0000 \\
\textbf{SWU} & 1.0000 & 1.0000 & 1.0000 & 1.0000 & 0.9648 & 1.0000 & 1.0000 & 1.0000 & -- \\
\midrule
\multicolumn{10}{c}{\textbf{Brier} $\downarrow$} \\
\midrule
\textbf{DE} & -- & 0.0664 & 0.0547 & 0.9609 & 0.9609 & 0.6445 & 0.1367 & 1.0000 & 0.9609 \\
\textbf{TTA} & 0.0664 & -- & 0.9160 & 0.0664 & 0.6445 & 0.0879 & 0.0547 & 0.3418 & 0.0547 \\
\textbf{MCDO} & 0.0547 & 0.9160 & -- & 0.0547 & 0.1777 & 0.0547 & 0.0547 & 0.0879 & 0.0547 \\
\textbf{TS} & 0.9609 & 0.0664 & 0.0547 & -- & 0.8438 & 0.9609 & 0.0547 & 1.0000 & 0.0547 \\
\textbf{DUQ} & 0.9609 & 0.6445 & 0.1777 & 0.8438 & -- & 0.9609 & 0.1777 & 1.0000 & 0.9609 \\
\textbf{DDU-Seg} & 0.6445 & 0.0879 & 0.0547 & 0.9609 & 0.9609 & -- & 0.0547 & 1.0000 & 0.4082 \\
\textbf{DUE} & 0.1367 & 0.0547 & 0.0547 & 0.0547 & 0.1777 & 0.0547 & -- & 0.8477 & 0.0547 \\
\textbf{ConfidNet} & 1.0000 & 0.3418 & 0.0879 & 1.0000 & 1.0000 & 1.0000 & 0.8477 & -- & 1.0000 \\
\textbf{SWU} & 0.9609 & 0.0547 & 0.0547 & 0.0547 & 0.9609 & 0.4082 & 0.0547 & 1.0000 & -- \\
\midrule
\multicolumn{10}{c}{\textbf{AUROC} $\uparrow$} \\
\midrule
\textbf{DE} & -- & 0.0742 & 0.0547 & 0.0742 & 1.0000 & 0.0742 & 1.0000 & 1.0000 & 1.0000 \\
\textbf{TTA} & 0.0742 & -- & 0.0547 & 0.7559 & 0.4453 & 0.6445 & 0.7559 & 1.0000 & 0.4453 \\
\textbf{MCDO} & 0.0547 & 0.0547 & -- & 0.0547 & 0.0547 & 0.0547 & 0.0547 & 0.0879 & 0.0547 \\
\textbf{TS} & 0.0742 & 0.7559 & 0.0547 & -- & 0.0742 & 0.9160 & 0.3555 & 1.0000 & 0.0547 \\
\textbf{DUQ} & 1.0000 & 0.4453 & 0.0547 & 0.0742 & -- & 0.0742 & 1.0000 & 1.0000 & 1.0000 \\
\textbf{DDU-Seg} & 0.0742 & 0.6445 & 0.0547 & 0.9160 & 0.0742 & -- & 0.2734 & 1.0000 & 0.0547 \\
\textbf{DUE} & 1.0000 & 0.7559 & 0.0547 & 0.3555 & 1.0000 & 0.2734 & -- & 1.0000 & 1.0000 \\
\textbf{ConfidNet} & 1.0000 & 1.0000 & 0.0879 & 1.0000 & 1.0000 & 1.0000 & 1.0000 & -- & 0.1250 \\
\textbf{SWU} & 1.0000 & 0.4453 & 0.0547 & 0.0547 & 1.0000 & 0.0547 & 1.0000 & 0.1250 & -- \\
\midrule
\multicolumn{10}{c}{\textbf{AURC} $\downarrow$} \\
\midrule
\textbf{DE} & -- & 0.7090 & 0.0547 & 0.0547 & 1.0000 & 0.0547 & 1.0000 & 1.0000 & 1.0000 \\
\textbf{TTA} & 0.7090 & -- & 0.0703 & 1.0000 & 0.5859 & 1.0000 & 0.8398 & 1.0000 & 0.0703 \\
\textbf{MCDO} & 0.0547 & 0.0703 & -- & 0.0547 & 0.0547 & 0.0547 & 0.0547 & 0.0879 & 0.0547 \\
\textbf{TS} & 0.0547 & 1.0000 & 0.0547 & -- & 0.0703 & 1.0000 & 0.3828 & 0.1367 & 0.0547 \\
\textbf{DUQ} & 1.0000 & 0.5859 & 0.0547 & 0.0703 & -- & 0.0703 & 1.0000 & 0.2246 & 1.0000 \\
\textbf{DDU-Seg} & 0.0547 & 1.0000 & 0.0547 & 1.0000 & 0.0703 & -- & 0.3828 & 0.1367 & 0.0547 \\
\textbf{DUE} & 1.0000 & 0.8398 & 0.0547 & 0.3828 & 1.0000 & 0.3828 & -- & 0.2246 & 1.0000 \\
\textbf{ConfidNet} & 1.0000 & 1.0000 & 0.0879 & 0.1367 & 0.2246 & 0.1367 & 0.2246 & -- & 0.2246 \\
\textbf{SWU} & 1.0000 & 0.0703 & 0.0547 & 0.0547 & 1.0000 & 0.0547 & 1.0000 & 0.2246 & -- \\
\bottomrule
\end{tabular*}
\caption{Holm-corrected pairwise Wilcoxon $p$-values for LiTS across all metrics. Off-diagonal entries compare row and column methods across test cases; the matrix is symmetric, and \verySmallP{} denotes $p<0.0001$.}
\label{tab:pvalues_lits}
\end{table*}

\subsection{Ablations}

The ACDC ablations, detailed in the supplementary PDF, support three design choices. Because the backbone is frozen, Dice remains $0.9035{\pm}0.0044$ for all variants. Calibration and ranking should be separated: calibration-only keeps Brier competitive at $0.0114$ but collapses AUROC/AURC to $0.7078/28.0538$, while ranking-only recovers $0.9824/2.9853$ but worsens Brier to $0.0122$. Ranking losses are central. Removing NLL, error-correlation, pairwise, tail, and trust terms gives AURC $4.4299$, $4.5017$, $16.7523$, $3.4856$, and $2.7331$, respectively, compared with $2.4885$ for the full model. The pairwise term is especially important, dropping AUROC from $0.9838$ to $0.9275$ when removed. Finally, \method{} ablations using a direct uncertainty head, fixed perturbation scales, or removing the aleatoric branch yield AURC $4.0557$, $3.0440$, and $4.8666$, respectively. These drops indicate that the learned perturbation-probe mechanism contributes beyond a generic auxiliary uncertainty head.

\subsection{Efficiency and Selective-Prediction Behavior}

\method{} adds about $0.1$M parameters and preserves single-backbone-pass inference. Measured wall-clock times for \method{} are $0.23/0.54/14.46$ s on ACDC/BraTS2024/LiTS, compared with $1.72/2.63/39.48$ s for Deep Ensembles, $0.50/1.18/31.31$ s for TTA, $1.25/2.96/90.38$ s for MC Dropout, $0.92/3.24/45.52$ s for DUE, and $1.59/4.79/60.97$ s for ConfidNet. Thus the main comparison is not only accuracy but whether a lightweight add-on can recover much of the ranking quality of expensive uncertainty references.

The risk--coverage examples in the supplement show that \method{} is usually in the favorable low-risk group as uncertain voxels are rejected, although the leading curve can vary by case and coverage level. In the accuracy-threshold plots, retained voxels generally become more accurate as the confidence cutoff increases. This supports selective acceptance and targeted review, rather than treating uncertainty only as a scalar calibration diagnostic.

\section{Discussion and Limitations}
\label{sec:discussion}

\method{} adds supervised reliability estimation to a frozen segmentor without changing hard predictions or rerunning the heavy 3D backbone. This is most useful for selective prediction, where AURC measures whether confident voxels can be retained while uncertain ones are escalated for review.

We evaluate public benchmarks rather than held-out site, scanner, vendor, or protocol shifts. BraTS2024 and LiTS are multi-institutional or multi-center datasets, and ACDC includes disease-group and acquisition variation, but our splits do not isolate those factors. \method{} also requires labeled data for the uncertainty head, so deployment should reserve a calibration/validation set. Within this scope, \method{} combines frozen-backbone deployment, explicit calibration/ranking separation, and direct voxel-wise risk-ordering losses.

\clearpage
\bibliographystyle{splncs04}
\bibliography{main}

@article{Bernard2018DeepLearningCardiacMRI,
  title   = {Deep Learning Techniques for Automatic MRI Cardiac Multi-Structures Segmentation and Diagnosis: Is the Problem Solved?},
  author  = {Bernard, Olivier and others},
  journal = {IEEE Trans. Med. Imaging},
  volume  = {37},
  number  = {11},
  pages   = {2514--2525},
  year    = {2018}
}

@article{Bilic2019LiTS,
  title   = {The Liver Tumor Segmentation Benchmark (LiTS)},
  author  = {Bili{\'c}, Patrick and others},
  journal = {Med. Image Anal.},
  volume  = {84},
  pages   = {102680},
  year    = {2023}
}

@misc{Cardoso2022MONAI,
  title        = {MONAI: An Open-Source Framework for Deep Learning in Healthcare},
  author       = {Cardoso, M. Jorge and others},
  year         = {2022},
  eprint       = {2211.02701},
  archivePrefix = {arXiv}
}

@misc{deverdier20242024braintumorsegmentation,
  title        = {The 2024 Brain Tumor Segmentation (BraTS) Challenge: Glioma Segmentation on Post-treatment MRI},
  author       = {Correia de Verdier, Maria and others},
  year         = {2024},
  eprint       = {2405.18368},
  archivePrefix = {arXiv}
}

@inproceedings{Gal2016DropoutBayes,
  title     = {Dropout as a Bayesian Approximation: Representing Model Uncertainty in Deep Learning},
  author    = {Gal, Yarin and Ghahramani, Zoubin},
  booktitle = {ICML},
  year      = {2016}
}

@inproceedings{Goodfellow2015Explaining,
  title     = {Explaining and Harnessing Adversarial Examples},
  author    = {Goodfellow, Ian J. and Shlens, Jonathon and Szegedy, Christian},
  booktitle = {ICLR},
  year      = {2015}
}

@inproceedings{Guo2017Calibration,
  title     = {On Calibration of Modern Neural Networks},
  author    = {Guo, Chuan and Pleiss, Geoff and Sun, Yu and Weinberger, Kilian Q.},
  booktitle = {ICML},
  year      = {2017}
}

@inproceedings{Lakshminarayanan2017DeepEnsembles,
  title     = {Simple and Scalable Predictive Uncertainty Estimation Using Deep Ensembles},
  author    = {Lakshminarayanan, Balaji and Pritzel, Alexander and Blundell, Charles},
  booktitle = {NeurIPS},
  year      = {2017}
}

@inproceedings{Liu2020DUE,
  title     = {Simple and Principled Uncertainty Estimation with Deterministic Deep Learning via Distance Awareness},
  author    = {Liu, Jeremiah Zhe and Lin, Zi and Padhy, Shreyas and Tran, Dustin and Bedrax-Weiss, Tania and Lakshminarayanan, Balaji},
  booktitle = {NeurIPS},
  year      = {2020}
}

@misc{Mukhoti2021DDUSeg,
  title        = {Deep Deterministic Uncertainty for Semantic Segmentation},
  author       = {Mukhoti, Jishnu and van Amersfoort, Joost and Torr, Philip H. S. and Gal, Yarin},
  year         = {2021},
  eprint       = {2111.00079},
  archivePrefix = {arXiv}
}

@inproceedings{Mukhoti2023DDU,
  title     = {Deep Deterministic Uncertainty: A New Simple Baseline},
  author    = {Mukhoti, Jishnu and Kirsch, Andreas and van Amersfoort, Joost and Torr, Philip H. S. and Gal, Yarin},
  booktitle = {CVPR},
  pages     = {24384--24394},
  year      = {2023}
}

@inproceedings{VanAmersfoort2020DUQ,
  title     = {Uncertainty Estimation Using a Single Deep Deterministic Neural Network},
  author    = {van Amersfoort, Joost and Smith, Lewis and Teh, Yee Whye and Gal, Yarin},
  booktitle = {ICML},
  pages     = {9690--9700},
  year      = {2020}
}

@article{Wang2019TTA,
  title   = {Aleatoric Uncertainty Estimation with Test-Time Augmentation for Medical Image Segmentation with Convolutional Neural Networks},
  author  = {Wang, Guotai and Li, Wenqi and Aertsen, Michael and Deprest, Jan and Ourselin, Sebastien and Vercauteren, Tom},
  journal = {Neurocomputing},
  volume  = {338},
  pages   = {34--45},
  year    = {2019}
}

@incollection{NIPS2019_8556,
   title = {Addressing Failure Prediction by Learning Model Confidence},
   author = {Corbi\`{e}re, Charles and THOME, Nicolas and Bar-Hen, Avner and Cord, Matthieu and P\'{e}rez, Patrick},
   booktitle = {Advances in Neural Information Processing Systems 32},
   editor = {H. Wallach and H. Larochelle and A. Beygelzimer and F. d\textquotesingle Alch\'{e}-Buc and E. Fox and R. Garnett},
   pages = {2902--2913},
   year = {2019},
   publisher = {Curran Associates, Inc.},
   url = {http://papers.nips.cc/paper/8556-addressing-failure-prediction-by-learning-model-confidence.pdf}
}

\clearpage
\begingroup
\newgeometry{margin=0.42in}
\pagestyle{empty}
\def\AppendixInMain{1}
\ifdefined\AppendixInMain
\let\FinishAppendix\relax
\else
\documentclass[10pt]{article}
\usepackage[margin=0.42in]{geometry}
\usepackage[T1]{fontenc}
\usepackage{amsmath,amssymb}
\usepackage{booktabs}
\usepackage{array}
\usepackage{graphicx}
\usepackage{multicol}

\newcommand{\method}{SegWithU}
\newcommand{\verySmallP}{\ensuremath{\ll}}

\begin{document}
\def\FinishAppendix{\end{document}}
\fi
\pagestyle{empty}
\setlength{\parindent}{0pt}
\setlength{\parskip}{0.8pt}
\setlength{\columnsep}{0.15in}
\setlength{\tabcolsep}{2.0pt}
\renewcommand{\arraystretch}{0.75}
\begin{center}
{\large\bf Supplementary Appendix: \method}\\[-1pt]
{\small Anonymous Authors}
\end{center}
\vspace{-6pt}
\fontsize{8}{9.5}\selectfont
\sloppy

\begin{multicols}{2}
\section*{A. Formula Reference}
\textbf{Backbone/probes.}
\(z=f_\theta(x)\), \(r_m=\mathrm{Resize}(W_mh^{(m)})\), \(h=\phi_{\rm fuse}([r_1,\ldots,r_M])\), \(v=\psi(h)\), \(\sigma_r=\mathrm{softplus}(\alpha_r)+\epsilon\).
For zero-indexed \(k,r\), \(q_{kr}=\cos\{\pi(k+1)(r+0.5)/R\}\), \(u^{(k)}_r=+1\) if \(q_{kr}\ge0\) and \(-1\) otherwise, \(\Delta z(u^{(k)})=A((\sigma\odot u^{(k)})\odot v)\), \(p^{(k)}=\mathrm{softmax}(z+\Delta z(u^{(k)}))\).
\(U_{\rm epi}=\sum_c\mathrm{Var}_k(p^{(k)}_c)\), \(U_{\rm probe}=R^{-1}\sum_r v_r^2\), \(U_{\rm res}=C^{-1}\sum_c\Delta z_c^2\).

\textbf{Calibration/ranking.}
\(p=\mathrm{softmax}(z)\), \(m=p_{(1)}-p_{(2)}\), \(w=\exp(-\gamma m)\), \(U_{\rm ale}=\mathrm{softplus}(\psi_{\rm ale}(h))\).
\[
\begin{aligned}
U_{\rm cal}&=\mathrm{softplus}\{\psi_{\rm cal}[
\log(1+U_{\rm epi}+U_{\rm res}),\\[-2pt]
&\hspace{22pt}\log(1+U_{\rm ale}),m]\},\quad
\tilde z=z/\sqrt{1+U_{\rm cal}},\\[-1pt]
U_{\rm anchor}&=\log(1+U_{\rm epi})+\tfrac12\log(1+U_{\rm res})
+\tfrac14\log(1+U_{\rm cal})\\[-2pt]
&\hspace{22pt}+\tfrac14 H(p)/\log C+w,\\[-1pt]
U_{\rm rnk}&=(1+0.1\tanh a)U_{\rm anchor}+b+\mathrm{softplus}(c)w .
\end{aligned}
\]
\textbf{Objective.}
\(\mathcal L=0.5\mathcal L_{\rm NLL}+0.25(\mathcal L_{\rm EC}+\mathcal L_{\rm Pairwise}+\mathcal L_{\rm Tail})+0.05(\mathcal L_{\rm Trust}+\mathcal L_{\rm Anchor}+\mathcal L_{\rm Residual})\), with \(e_i=\mathbf1[\arg\max_c z_{i,c}\ne y_i]\) and \(u_i=U_{{\rm rnk},i}\).

\textbf{Loss definitions.}
Let \(\hat y_i=\arg\max_c z_{i,c}\), \(e_i=\mathbf1[\hat y_i\ne y_i]\), \(u_i=U_{{\rm rnk},i}\), \(\bar u_i=(u_i-\mu_u)/(\sigma_u+\epsilon)\), \(\tilde p_i=\mathrm{softmax}(\tilde z_i)\), \(p=\mathrm{softmax}(z)\), \(p^\Delta=\mathrm{softmax}(z+\Delta z)\), \(\bar a_i=(U_{{\rm anchor},i}-\mu_a)/(\sigma_a+\epsilon)\), voxels \(\mathcal V\), and sampled error/correct pairs \(\mathcal P\). Then \( \mathcal L_{\rm NLL}=-|\mathcal V|^{-1}\sum_i\log\tilde p_{i,y_i}\); \( \mathcal L_{\rm EC}=|\mathcal V|^{-1}\sum_i{\rm BCE}(\bar u_i/\tau,e_i)\); \( \mathcal L_{\rm Pairwise}=|\mathcal P|^{-1}\sum_{(i,j)\in\mathcal P}{\rm softplus}((\bar u_j-\bar u_i+\delta)/\tau)\); \(q_i=\exp(-u_i/T)/\sum_j\exp(-u_j/T)\), \( \mathcal L_{\rm Tail}=\sum_iq_ie_i\); \( \mathcal L_{\rm Trust}=(|\mathcal V|C)^{-1}\sum_{i,c}(\Delta z_{i,c})^2+(4|\mathcal V|C)^{-1}\sum_{i,c}(p^\Delta_{i,c}-p_{i,c})^2\); \( \mathcal L_{\rm Anchor}=|\mathcal V|^{-1}\sum_i{\rm SmoothL1}(\bar u_i,\bar a_i)\); and \( \mathcal L_{\rm Residual}=|\mathcal V|^{-1}\sum_i(1-w_i)U_{{\rm res},i}\). Here BCE denotes BCEWithLogits.

\section*{B. Experimental Protocol}
\textbf{Datasets and splits.}
ACDC: 3D cardiac MRI, 4 classes, official 200/100 train/test split, resampled, patch $64{\times}128{\times}128$.
BraTS2024: 4-channel brain MRI, 5 classes, 200 train and 100 held-out cases sampled from the public training set, patch $128^3$.
LiTS: abdominal CT, 3 classes, final 10 cases held out, patch $128^3$.
Foreground patches are oversampled at 33\%. ACDC is resampled because through-plane spacing is coarse; BraTS and LiTS use no spacing alignment in the reported implementation. All datasets use dataset-specific intensity normalization.

\textbf{Backbones and inference.}
All backbones are 3D DynUNet models trained on five non-overlapping folds plus an all-data model. Fold/all Dice: ACDC $88.21$--$90.51/90.35\%$; BraTS2024 $55.87$--$60.07/62.75\%$; LiTS $71.77$--$76.75/78.21\%$.
Backbone training uses Dice+CE, SGD momentum $0.9$, LR $0.01$, polynomial decay $p=0.9$, batch sizes $16/8/8$, and 200 epochs.
All 3D inference uses sliding windows with Gaussian blending, overlap $0.5$, window batch size 2, and the dataset ROI above.

\textbf{Baselines.}
Deep Ensembles average the five fold backbones and score uncertainty from predictive-distribution mutual information.
TTA uses the 8 axis-aligned 3D flip combinations, inverse-flips logits, averages probabilities, and scores entropy/variance.
MC Dropout enables only dropout at inference on a dropout-enabled DynUNet and averages repeated stochastic sliding-window predictions.
Temperature Scaling fits one scalar temperature on validation data by NLL; fitted temperatures are close to $1.5$, and entropy of tempered probabilities is the ranking score.
DUQ uses BCE plus gradient penalty $\lambda=0.5$, SGD momentum $0.9$, LR $0.05$, MultiStep decay, centroid size 8, length scale 0.1, and EMA $\gamma=0.999$.
DDU-Seg trains Dice+CE, then fits class-conditional Gaussian densities on penultimate features using streaming Chan covariance estimates.
DUE uses an SVGP head with RBF kernel, $M=16$ inducing points for ACDC and $M=64$ for BraTS/LiTS, K-means initialization, auxiliary CE weight $0.5$, spectral normalization, 32 posterior samples per spatial location, and spatial subsampling to 32,768 locations per step.
ConfidNet learns a confidence head that regresses the model's true-class probability (TCP) from intermediate features of the DynUNet.

\textbf{\method{} implementation.}
\method{} trains only a frozen-backbone uncertainty head. Multi-scale taps are \texttt{upsamples.1.conv\_block}, \texttt{upsamples.2.conv\_block}, and \texttt{output\_block}, with channels $(256,128,32)$ fused to 32 channels.
Default probes use $R=K=8$, $\sigma_{\rm init}=0.1$, margin weight $\gamma=4$, and the aleatoric branch enabled.
Optimization uses AdamW, LR $10^{-3}$, $\beta=(0.9,0.999)$, $\epsilon=10^{-8}$, cosine annealing to $3{\times}10^{-4}$, gradient clipping at 12, up to 200 epochs, early-stopping tolerance 10, and no deep supervision.
Loss weights are
$\lambda_{\rm nll}=0.5$,
$\lambda_{\rm ec}=\lambda_{\rm pair}=\lambda_{\rm tail}=0.25$,
$\lambda_{\rm trust}=\lambda_{\rm anchor}=\lambda_{\rm res}=0.05$,
and $\lambda_{\rm seg}=0$.

\textbf{Evaluation details.}
Ranking maps are: ensemble mutual information for Deep Ensembles; predictive entropy for MC Dropout, TTA, and TS; native aleatoric/deterministic scores for DUQ, DDU-Seg, and DUE; $U_{\rm rnk}$ for \method. $1-\widehat{\mathrm{TCP}}$ for ConfidNet, where its auxiliary confidence head approximates the voxel-wise true-class probability. \method{} uses $U_{\rm cal}$ only for probability tempering in Brier/NLL-style scores. AURC sorts voxels by increasing uncertainty and numerically integrates risk--coverage; risk--coverage plots also use random-rejection and oracle references. Accuracy-threshold plots use 101 thresholds on $[0,1]$.

\textbf{Comparison caveats.}
DUQ, DDU, and DUE were originally classification methods; dense segmentation treats voxels as conditionally independent samples and changes computational scaling.
ConfidNet has an optional stage that fine-tunes a copy of the segmentation model after training the confidence network. We omit this fine-tuning stage to maintain the same frozen-backbone constraint as \method{}; additionally, in our experiments, the fine-tuned variant did not improve any metric on ACDC, BraTS, or LiTS compared to the original.
Original baselines did not use MONAI DynUNet, so absolute values may differ from their papers. DUQ hyperparameters were inherited from low-resolution classification; Deep Ensembles omit adversarial training; MC Dropout depends on where dropout appears in DynUNet; DUE inference is stochastic unless the GP sampling seed is fixed.

\section*{C. ACDC Ablations}
\begin{center}
\scriptsize
\begin{tabular}{llccc}
\toprule
Group & Variant & Brier $\downarrow$ & AUROC $\uparrow$ & AURC $\downarrow$ \\
\midrule
Map & Cal. only & $.0114{\pm}.0006$ & $.7078{\pm}.0050$ & $28.0538{\pm}1.4905$ \\
Map & Rank only & $.0122{\pm}.0006$ & $.9824{\pm}.0026$ & $2.9853{\pm}.7933$ \\
Loss & No NLL & $.0113{\pm}.0006$ & $.9782{\pm}.0035$ & $4.4299{\pm}1.3205$ \\
Loss & No EC & $.0113{\pm}.0006$ & $.9746{\pm}.0034$ & $4.5017{\pm}1.1033$ \\
Loss & No pair & $.0116{\pm}.0006$ & $.9275{\pm}.0062$ & $16.7523{\pm}2.4787$ \\
Loss & No tail & $.0114{\pm}.0006$ & $.9795{\pm}.0031$ & $3.4856{\pm}.9728$ \\
Loss & No trust & $.0113{\pm}.0006$ & $.9823{\pm}.0024$ & $2.7331{\pm}.6607$ \\
Probe & Direct & $.0114{\pm}.0006$ & $.9768{\pm}.0031$ & $4.0557{\pm}.9928$ \\
Probe & Fixed $\sigma$ & $.0113{\pm}.0006$ & $.9813{\pm}.0026$ & $3.0440{\pm}.7501$ \\
Probe & No ale. & $.0114{\pm}.0006$ & $.9731{\pm}.0038$ & $4.8666{\pm}1.2919$ \\
\midrule
Full & \method & $\mathbf{.0113{\pm}.0006}$ & $\mathbf{.9838{\pm}.0022}$ & $\mathbf{2.4885{\pm}.6077}$ \\
\bottomrule
\end{tabular}
\end{center}
Dice is $0.9035{\pm}0.0044$ for all rows because the backbone and hard segmentation are frozen. AURC is in $10^{-4}$ units. Calibration-only preserves Brier but fails ranking; ranking-only recovers ordering but worsens Brier; pairwise loss is the largest single loss contributor; direct/fixed/no-aleatoric variants show that the learned perturbation probe is not a generic auxiliary head.

\section*{D. Extra Sensitivity Ablations}
\begin{center}
\scriptsize
\begin{tabular}{lccc|lccc}
\toprule
$R$ & Brier & AUROC & AURC & $\gamma$ & Brier & AUROC & AURC\\
\midrule
4 & $.0113$ & $.9795$ & $3.6617$ & 1 & $.0113$ & $.9783$ & $3.7818$\\
16 & $.0113$ & $.9815$ & $3.0500$ & 2 & $.0113$ & $.9797$ & $3.8112$\\
32 & $.0113$ & $.9806$ & $3.3402$ & 8 & $.0114$ & $.9784$ & $3.9717$\\
\textbf{8} & $\mathbf{.0113}$ & $\mathbf{.9838}$ & $\mathbf{2.4885}$ & \textbf{4} & $\mathbf{.0113}$ & $\mathbf{.9838}$ & $\mathbf{2.4885}$\\
\bottomrule
\end{tabular}
\end{center}
Default $R=8$ and $\gamma=4$ give the best ACDC ranking quality, suggesting a compact-probe bias--variance tradeoff and a moderate ambiguity-emphasis optimum.
Single-tap vs multi-scale tapping: single late tap gives Brier/AUROC/AURC $0.0116/0.9743/3.9136$; multi-scale tapping improves to $0.0113/0.9838/2.4885$, indicating that uncertainty cues benefit from both coarse semantic and finer spatial features.

\section*{E. Statistical and Case-Level Details}
Pairwise statistics use per-case Wilcoxon signed-rank tests with Holm correction within each dataset--metric family. The score vectors below follow (DE, TTA, MCDO, TS, DUQ, DDU-Seg, DUE, ConverSeg, \method).
\begin{center}
\scriptsize
\begin{tabular}{lcc}
\toprule
Dataset & DE/TTA/MCDO/TS & DUQ/DDU/DUE/ConfidNet/\method\\
\midrule
ACDC & $+16,-20,-20,-1$ & $-15,-5,+14,+6,+25$\\
BraTS & $-2,+3,-29,-5$ & $+6,-6,+9,0,+24$\\
LiTS & $0,0,0,0$ & $0,0,0,0,0$\\
All & $+14,-17,-49,-6$ & $-9,-11,+23,+6,+49$\\
\bottomrule
\end{tabular}
\end{center}
LiTS is set to zero for corrected significance scoring because $n=10$ is underpowered. \method{} is the only method never significantly outperformed on ACDC/BraTS2024.

\textbf{Qualitative case metrics.}
\begin{center}
\scriptsize
\begin{tabular}{lcccc}
\toprule
Case & Dice & Brier & AUROC & AURC\\
\midrule
ACDC 7 & .8867 & .0087 & .9905 & .7100\\
BraTS 44 & .5554 & .0017 & .9974 & .0347\\
LiTS 8 & .9254 & .0222 & .9695 & 5.5866\\
\bottomrule
\end{tabular}
\end{center}
Qualitatively, \method{} usually keeps uncertainty attached to anatomy and produces fewer isolated high-uncertainty regions away from anatomy than Deep Ensembles or DUQ; DUE is often closest.

\textbf{Risk and confidence curves.}
Per-case risk--coverage curves show \method{} usually in the favorable low-risk group, though not always the single best curve at every coverage. It more reliably avoids early degradation than MC Dropout and often DDU/TTA. ConverSeg is competitive on ACDC 7 and BraTS2024 44, but \method{} provides better error ranking across all three selected cases with the clearest separation on LiTS 8. In the accuracy-threshold curves, retained voxels generally become more accurate as confidence thresholds rise, supporting selective acceptance and targeted review rather than only scalar calibration.

\textbf{LiTS per-case \method{} summary.}
\begin{center}
\scriptsize
\begin{tabular}{lcc}
\toprule
Metric & Mean$\pm$SD & Range\\
\midrule
Dice & $.7821{\pm}.0343$ & --\\
Brier & $.0067{\pm}.0020$ & --\\
AUROC & $.9925{\pm}.0025$ & $.9695$--$.9985$\\
AURC & $.8193{\pm}.5117$ & $.0136$--$5.5866$\\
\bottomrule
\end{tabular}
\end{center}
Volume 129 is hardest for uncertainty ranking despite high Dice, illustrating that segmentation overlap and uncertainty ordering are related but distinct targets.

\textbf{Runtime and efficiency.}
\begin{center}
\tiny
\begin{tabular}{lccc}
\toprule
Method & Pass & Time A/B/L\\
\midrule
\multicolumn{4}{l}{Hardware: RTX Pro 6000 GPU}\\
\midrule
DE & 5 & 1.72/2.63/39.48 \\
MCDO & 20 & 1.25/2.96/90.38 \\
TTA & 8 & 0.50/1.18/31.31 \\
TS & 1 & 0.13/0.30/7.83 \\
DUQ & 1 & 0.13/0.31/8.12 \\
DDU & 1 & 0.13/0.30/7.82 \\
DUE & 1 & 0.92/3.24/45.52 \\
ConfidNet & 1 & 0.28/0.64/18.41 \\
\method & 1 & 0.23/0.54/14.46 \\
\bottomrule
\end{tabular}
\end{center}
Time is measured in seconds. \method{} adds 0.1M parameters.
\end{multicols}

\vspace{-10pt}
\section*{F. Additional Visual Plots}
\vspace{3pt}
\begin{center}
\tiny
\includegraphics[width=.76\textwidth]{images/segmentation-visualization.png}\\[3pt]
Hard-mask slices on ACDC, BraTS2024, and LiTS.
\\[3pt]
\includegraphics[width=.235\textwidth]{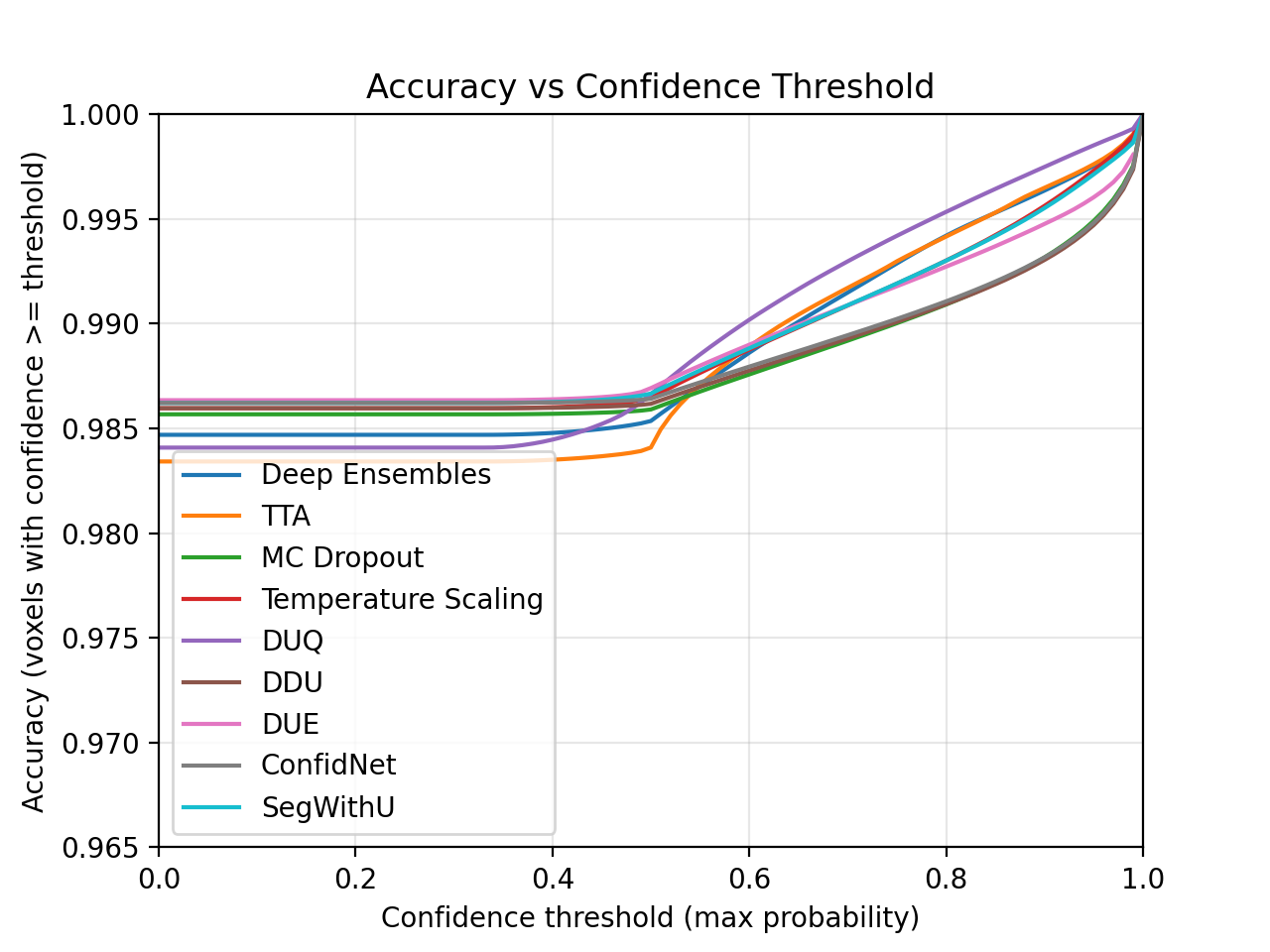}\hspace{2pt}
\includegraphics[width=.235\textwidth]{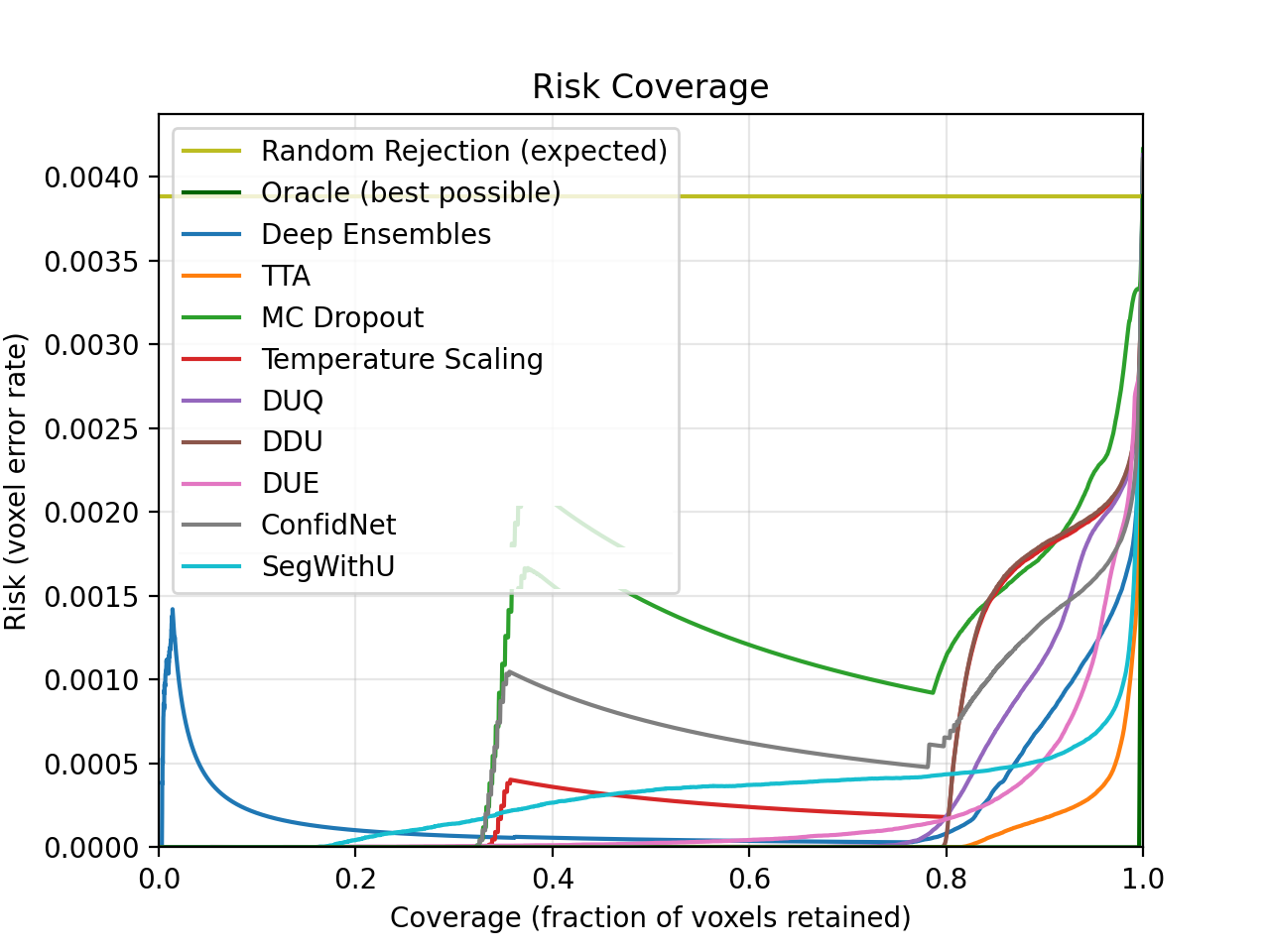}\hspace{2pt}
\includegraphics[width=.235\textwidth]{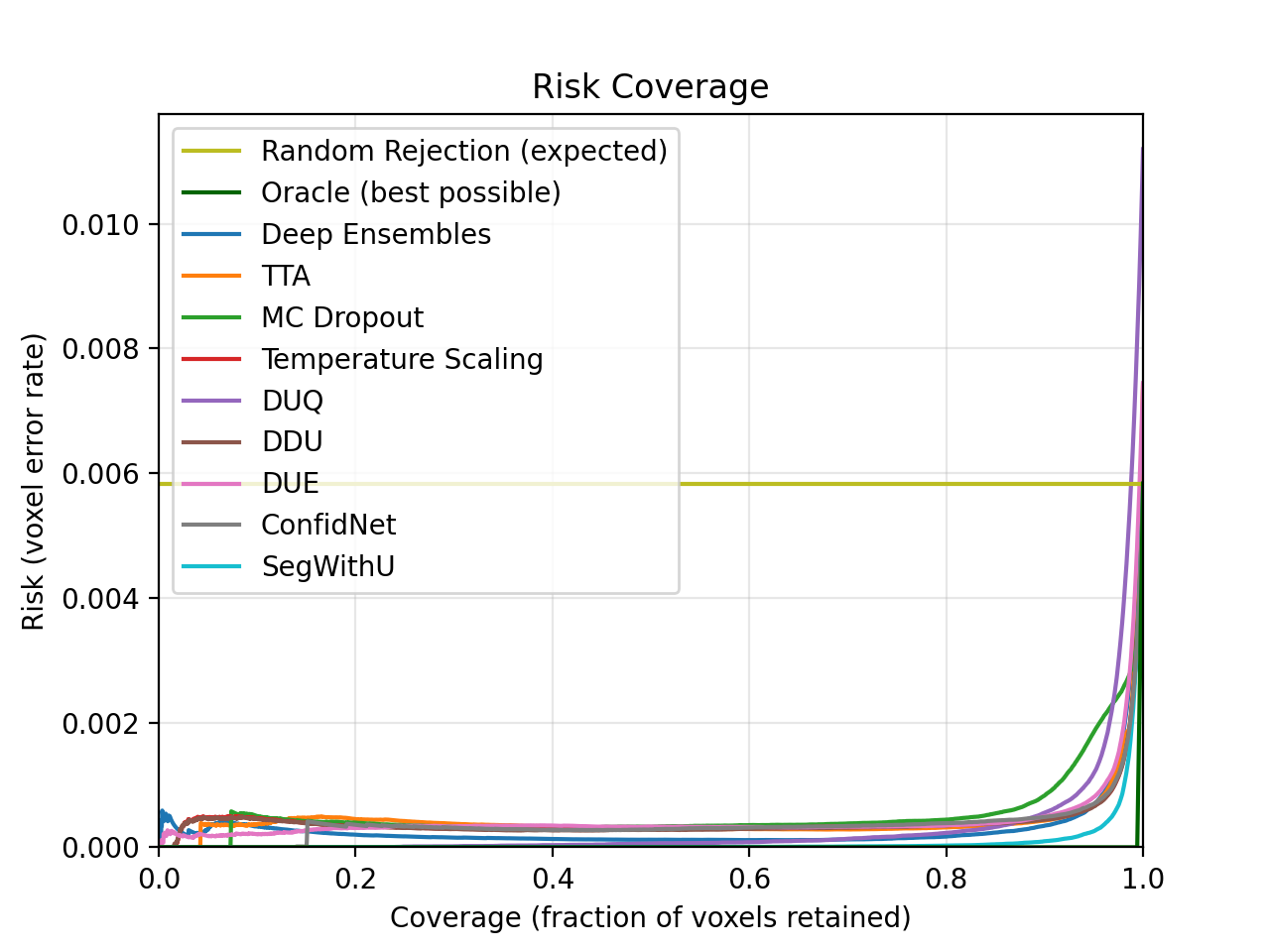}\\[-2pt]
Risk--coverage examples from ACDC, BraTS2024, and LiTS.\\[1pt]
\includegraphics[width=.235\textwidth]{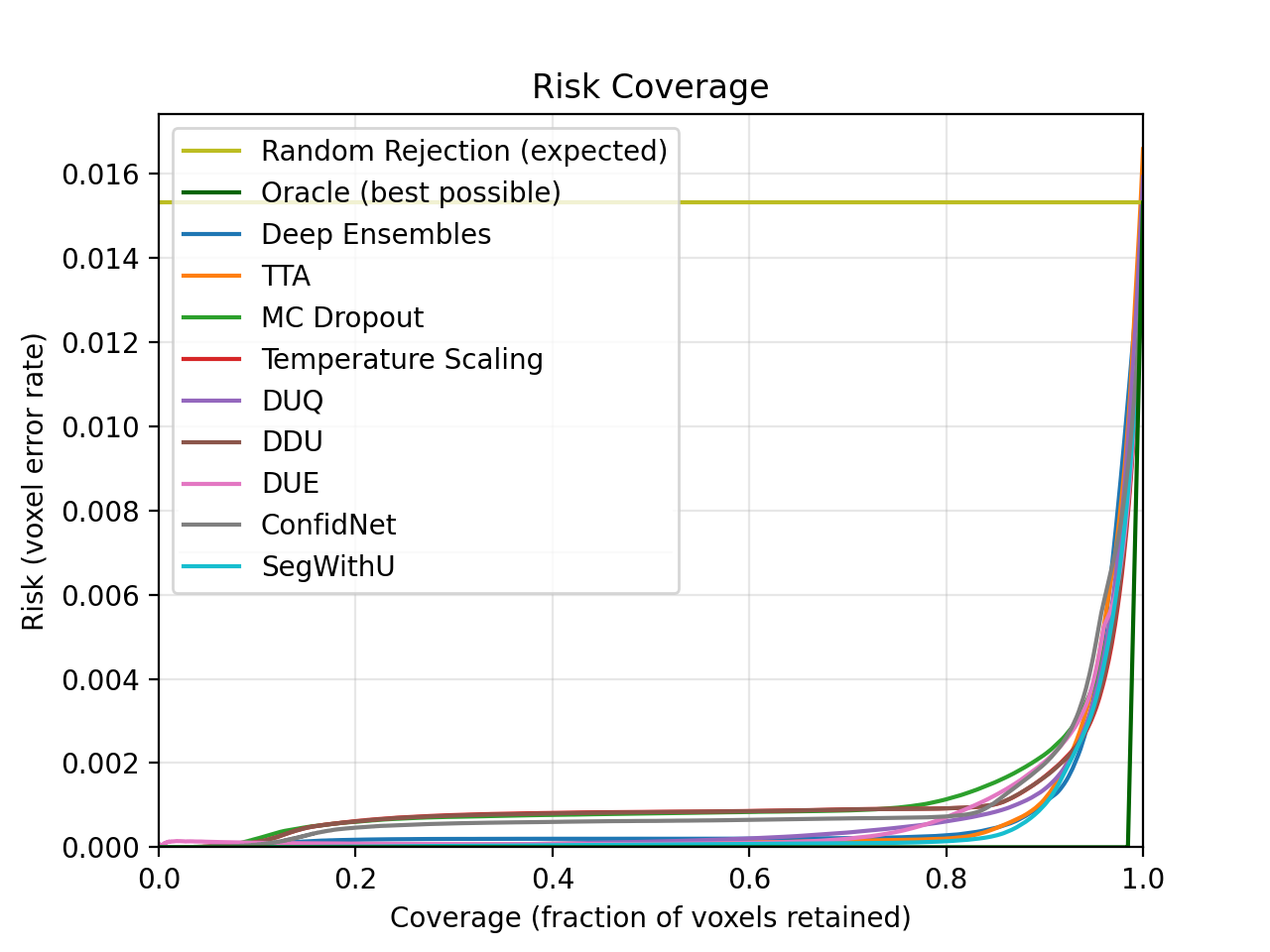}\hspace{2pt}
\includegraphics[width=.235\textwidth]{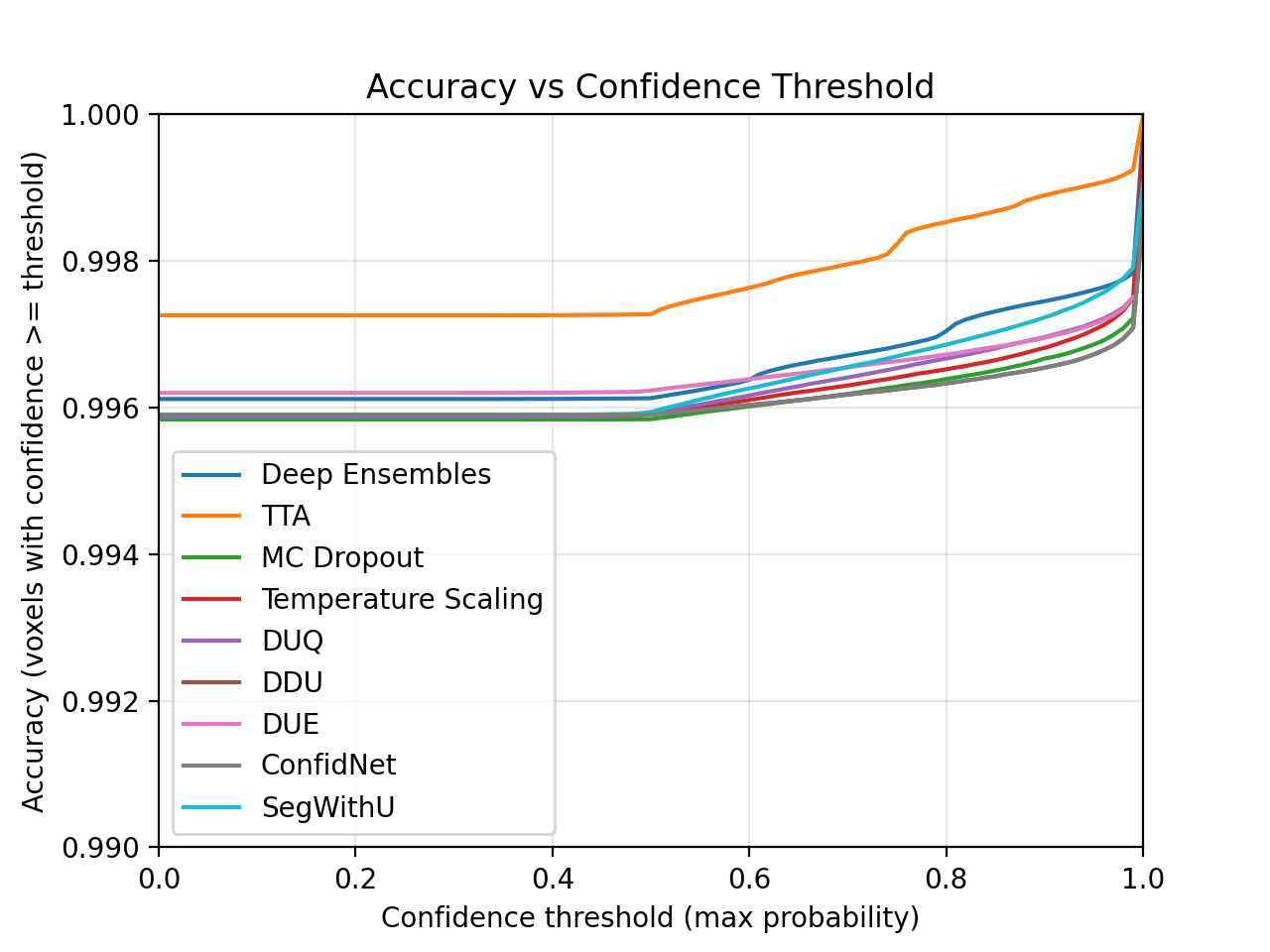}\hspace{2pt}
\includegraphics[width=.235\textwidth]{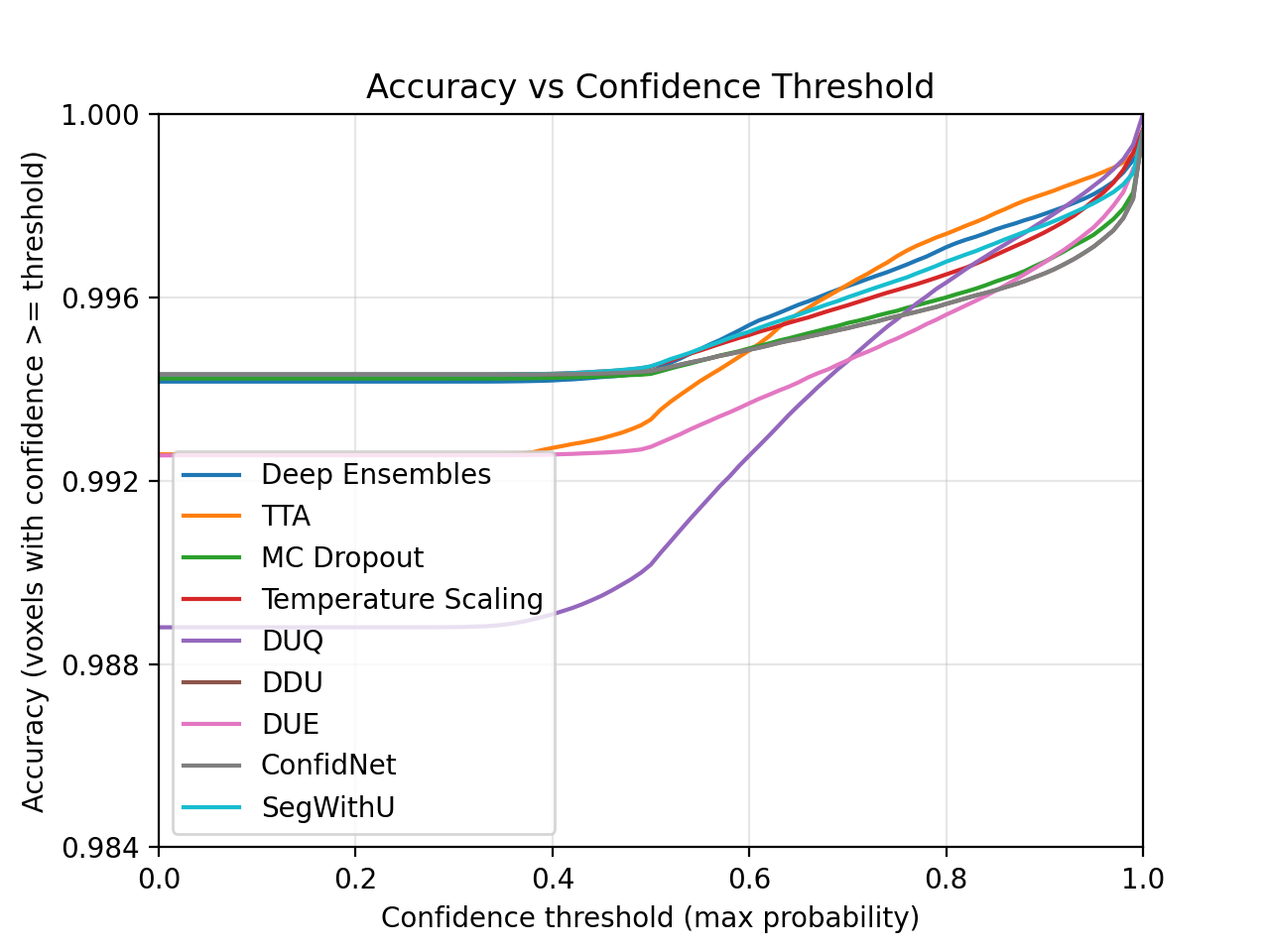}\\[-2pt]
Accuracy--threshold examples from the same selected cases.
\end{center}

\begin{minipage}{\textwidth}
\tiny
The mask slices show that \method{} preserves the frozen-backbone segmentation quality while adding reliability maps. Bar plots summarize aggregate metrics; risk--coverage and accuracy--threshold curves show selective-prediction behavior, with the leading method varying by case and coverage.
\end{minipage}

\FinishAppendix

\clearpage
\endgroup

\end{document}